\def\eg{{e.g. }}
\def\etal{{et al. }}
\def\etc{{etc. }}
\def\ie{{i.e. }}
\newcommand{\ignore}[1]{}
\documentclass[journal]{IEEEtran}
\ifCLASSINFOpdf
\else
\fi
\usepackage{url}
\usepackage{color}

\usepackage{times}
\usepackage{epsfig}
\def\baselinestretch{0.987}
\usepackage{graphicx,dblfloatfix}
\usepackage{amsmath}
\usepackage{amssymb}
\usepackage{comment}
\usepackage{booktabs}
\usepackage{caption}
\usepackage{subcaption}
\usepackage{array}
\graphicspath{{./figures/}}
\renewcommand{\baselinestretch}{0.97} 

\usepackage{expl3}
\ExplSyntaxOn
\newcommand\latinabbrev[1]{
  \peek_meaning:NTF . {
    #1\@}%
  { \peek_catcode:NTF a {
      #1.\@ }%
    {#1.\@}}}
\ExplSyntaxOff

\def\eg{\latinabbrev{e.g}}
\def\etal{\latinabbrev{et al}}
\def\etc{\latinabbrev{etc}}
\def\ie{\latinabbrev{i.e}}

\begin{document}
%


\title{Write a Classifier: Predicting Visual Classifiers from Unstructured Text}


%
%
%

\author{Mohamed ~Elhoseiny,~\IEEEmembership{Member,~IEEE,}
        Ahmed~Elgammal,~\IEEEmembership{Senior Member,~IEEE,}
        and~Babak~Saleh
}

\maketitle



%
\IEEEpeerreviewmaketitle

\begin{abstract}
People typically learn through exposure to visual concepts associated with linguistic descriptions. For instance, teaching visual object categories to children is often accompanied by descriptions in text or speech. In a machine learning context, these observations motivates us to ask whether this learning process could be computationally modeled to learn visual classifiers. More specifically, the main question of this work is how to utilize purely textual description of visual classes with no training images, to learn explicit visual classifiers for them. We propose and investigate two baseline formulations, based on regression and domain transfer, that predict a linear classifier. Then, we propose a new constrained optimization formulation that combines a regression function and a knowledge transfer function with additional constraints to predict the parameters of a linear classifier. We also propose a generic kernelized  models where a kernel classifier is predicted in the form defined by the representer theorem. The kernelized models allow defining and utilizing any two RKHS\footnote{Reproducing Kernel Hilbert Space} kernel functions in the visual space and text space, respectively. We finally propose a kernel function between unstructured text descriptions that builds on distributional semantics, which shows an advantage in our  setting  and could be useful for other applications. We applied all the studied models to predict visual classifiers on two fine-grained and challenging categorization datasets (CU Birds and Flower Datasets), and the results indicate successful predictions of our final model over several baselines that we designed.

\end{abstract}


\section{Introduction}
\label{intro}

One of the main challenges for scaling up object recognition systems is the lack of annotated images for real-world categories. 
Typically there are few images available for training classifiers for most of these categories. This is reflected in the number of images per category available for training in most object categorization datasets, which, as pointed out in~\cite{Salakhutdinov11}, shows a Zipf distribution. 
The problem of lack of training images becomes even more severe when we target recognition problems within a general category, \ie, fine-grained categorization, for example building classifiers for different bird species or flower types  (there are estimated over 10000 living bird species, similar for flowers). The largest bird image datasets contain only few hundred categories (e.g., CUBirds 200 dataset~\cite{wah2011caltech}). However,  descriptions about all the living birds are available in textual form (e.g.,~\cite{Avibase16,allaboutbirds16}).  
Researchers try to exploit shared knowledge between categories to target such scalability issue.
This motivated many researchers who looked into approaches that learn visual classifiers from few examples, \eg~\cite{deng2010does,fe2003bayesian,BartU05}.   This even motivated  more recent works on zero-shot learning of visual categories, where there are no training images available for test categories (unseen classes), \eg~\cite{Lampert09}. Such approaches exploit the similarity (visual or semantic) between seen classes and unseen ones, or describe unseen classes in terms of a learned vocabulary of semantic visual attributes.

  \begin{figure}[t!]
\centering
\vspace{-3mm}
    \includegraphics[width=0.83\linewidth]{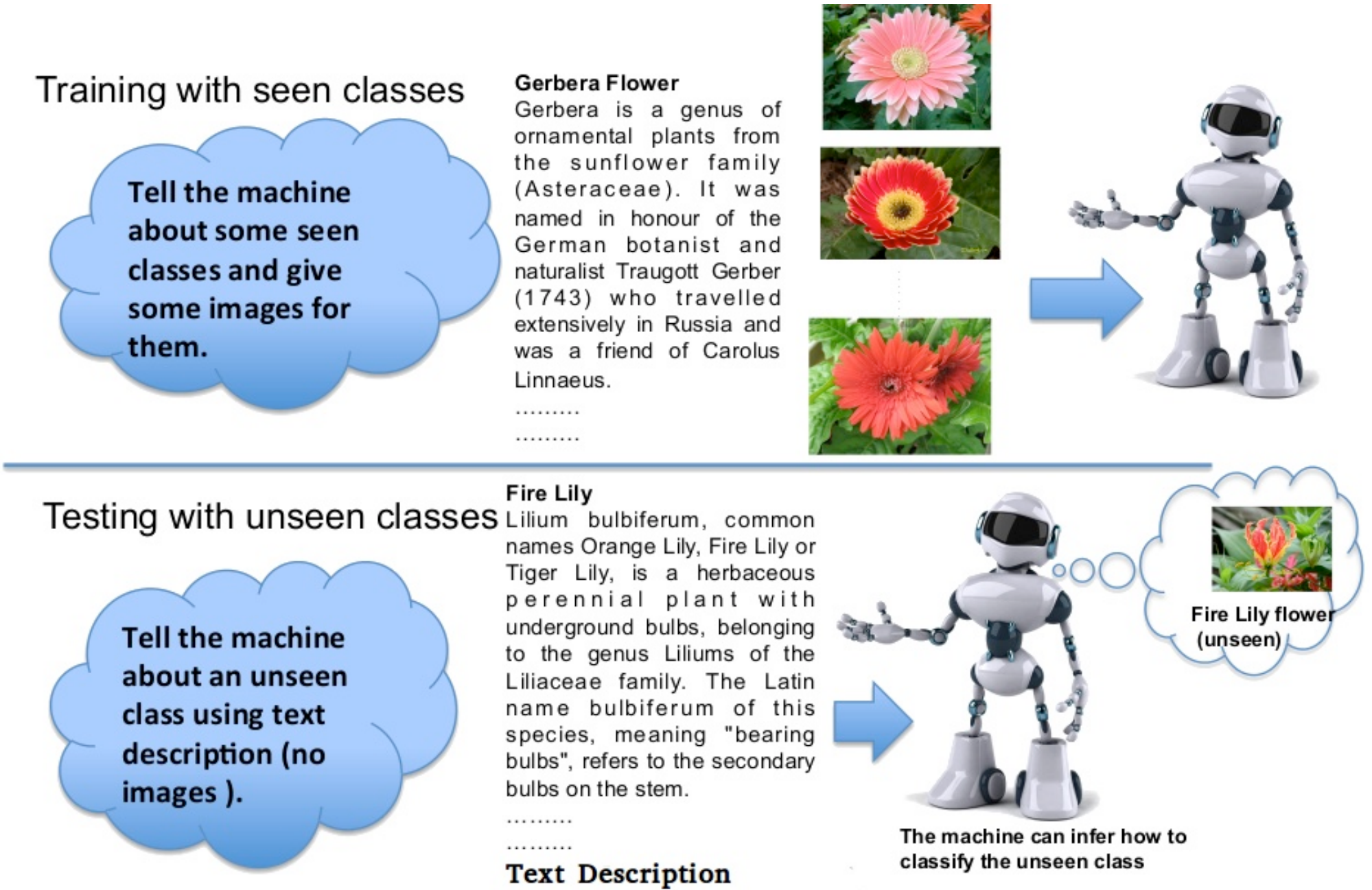} 
      \vspace{-2mm}
  \caption{{Our proposed setting where machine can predict unseen class from class-level unstructured  text description}}
  \label{fig:problem}
    \vspace{-6mm}
\end{figure}

In contrast to the lack of reasonably sized training sets for a large number of real world categories and subordinate categories, there are abundant of textual descriptions of these categories. This comes in the form of dictionary entries, encyclopedia articles, and various online resources. For example, it is possible to find several good descriptions of a ``bobolink'' in encyclopedias of birds, while there are only a few images available for that bird online. 

{\em The main question we address in this paper is how to use purely textual description of categories with no training images to learn visual classifiers for these categories.} In other words, we aim at zero-shot learning of object categories where the description of unseen categories comes in the form of typical text such as an encyclopedia entry; see Fig.~\ref{fig:problem}. We explicitly address the question of how to automatically decide which information to transfer between classes without the need of human intervention.  In contrast to most related work, we go beyond the simple use of tags and image captions, and apply standard Natural Language Processing techniques to typical text to learn visual classifiers.

Fine-grained categorization \ignore{(also known as subordinate categorization)} refers to classification of highly similar objects. This similarity can be due to natural intrinsic characteristics of subordinates of one category of objects (e.g. different breeds of dogs) or artificial subcategories of an object class (different types of airplanes). Diverse applications of fine-grained categories range from classification of natural species~\cite{wah2011caltech,Flower08,wang2009learning,liu2012dog} to retrieval of different types of commercial products~\cite{maji2013fine}. 
\begin{figure*}[t]
\centering
\input{bobolink_example}
\includegraphics[width=0.88\linewidth,height=.44\linewidth]{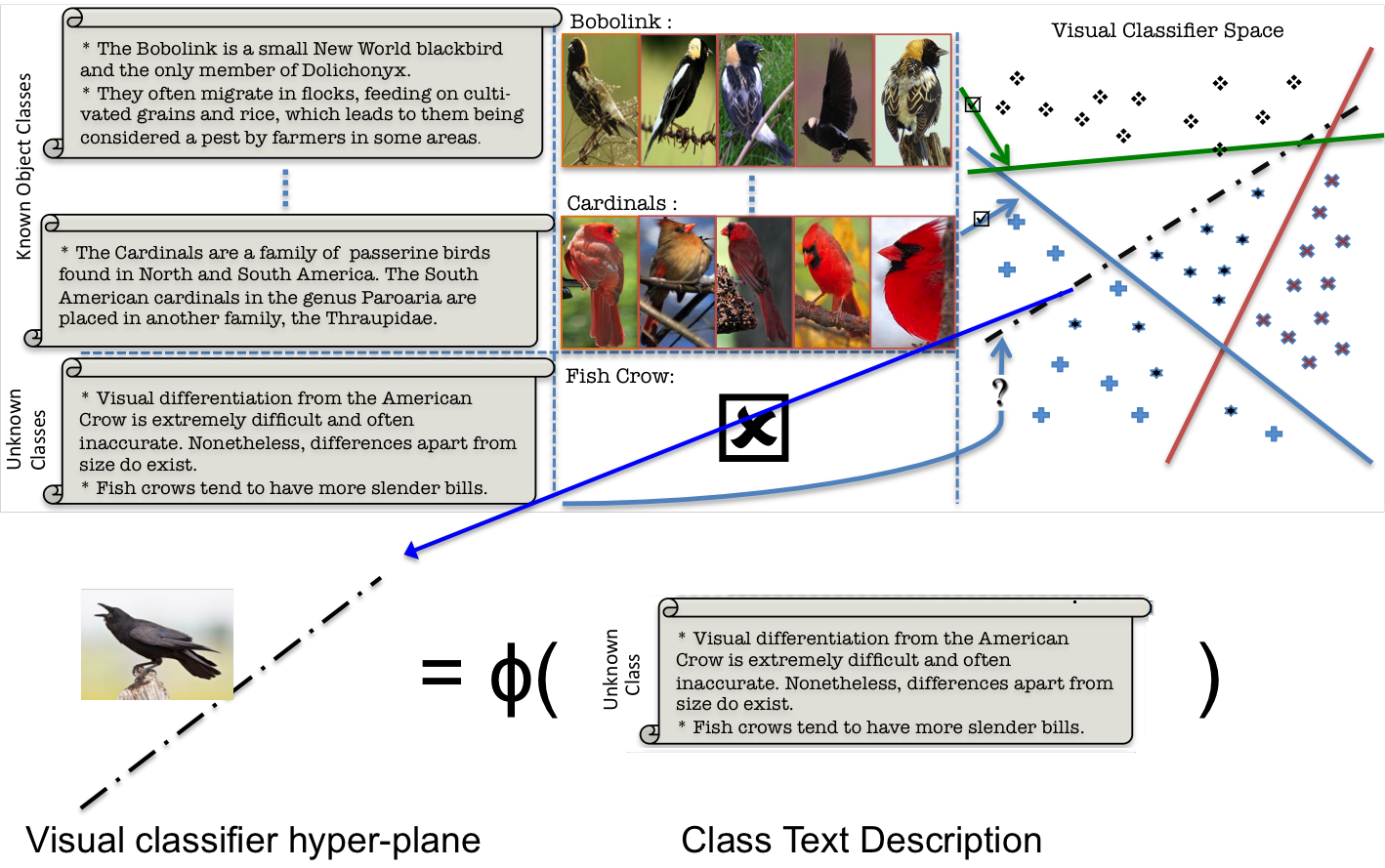}
\vspace{-2mm}
\caption{ Top: Example Wikipedia article about the Painted Bunting, with an example image. Bottom: The proposed learning setting. For each category we are give one (or more) textual description (only a synopsis of a larger text is shown),  and a set of training images. Our goal is to be able to predict a classifier for a  category based only on the narrative (zero-shot learning). }
\label{F:prob_def}
\vspace{-14pt}
\end{figure*}
In this problem, when we learn from an expert about different species of birds, the teacher will not just give you sample images of each species and their class labels; the teacher will tell you about discriminative visual or non-visual features for each species, similarities and differences between species, hierarchical relations between species, and many other aspects. The same learning experience takes place when you read a book or a web page to learn about  different species of birds; For example, Fig.~\ref{F:prob_def} shows an example narrative about the {\sf Bobolink}. Typically, the narrative tells you about the bird's taxonomy, highlights discriminative features about that bird and discusses similarities and differences between species, as well as within-species variations (male vs. female). The narrative might eventually show very few example images, which are often selected wisely to illustrate certain visual aspects that might be hard to explain in the narrative.  This learning strategy using textual narrative and images makes the learning effective without the huge number of images  that a typical visual learning algorithm would need to learn the class boundaries.

However, a narrative about a specific species does not contain only ``visually" relevant information, but also gives abundant information about the species's habitat, diet, mating habits, etc., which are not relevant for visual identification. In a sense, this information might be textual clutter for that task. The same problem takes place in images. While one image can be very effective in highlighting an important feature for learning, many images  might have a lot of visual clutter that makes their uses in learning not effective. 
Thus, a picture can be worth a thousand words, but not always, and an abundant number of pictures might not be the most effective way for learning. Similarly, one text paragraph can be worth a thousand pictures for learning a concept, but not always, and large amounts of text might not necessarily be effective.

The contribution of the paper is on exploring this new problem, which to the best of our knowledge, is firstly explored in the computer vision community in an earlier version of this work~\cite{Hoseini13}. We learn from an image corpus and a textual corpus, however not in the form of image-caption pairs, instead the only alignment between the corpora is at the level of the category. In particular, we address the problem of formulating a visual classifier prediction function ${\Phi (\cdot)}$, which predicts a  classifier of unseen visual class given its text description; see figure~\ref{F:prob_def}.  While a part of this work was  published in~\cite{Hoseini13}, we extend the work here to study more formulations to solve the problem in Sec.~\ref{formulation} (B,E). In addition, we propose a kernel method to explicitly  predict a kernel classifier  in the form defined in the representer theorem~\cite{rth01}.  The kernelized prediction has an advantage that it opens the door for using any kind of side information about classes, as long as kernels can be used on the side information representation.  The side information can be in the form of textual, parse trees, grammar, visual representations, concepts in the ontologies (adopted in NLP domain), or any form. We focus here on unstructured text descriptions.\ignore{; see figure~\ref{fig:problem}} The image features also do not need to be in a vectorized format. The kernelized classifiers also facilitate combining different types of features through a multi-kernel learning (MKL) paradigm, where the fusion of different features can be effectively achieved.

\ignore{We propose and investigate two baseline formulations based on regression and domain adaptation. Then we propose a new constrained optimization formulation that combines a regression function and a knowledge transfer function with additional constraints to solve the problem.} 

Beyond the introduction and the related work sections, the paper is structured as follows: Section~\ref{obvformulation} and ~\ref{sec_reganKT} details the problem definition and relation to regression and knowledge transfer models. Section~\ref{formulation} shows different formulations of $\Phi(\cdot)$ that we studied to predict a linear visual classifier; see figure~\ref{F:prob_def}. Section~\ref{sec:app} presents a kernelized version of our approach where  $\Phi(\cdot)$ predicts a kernel classifier in the form defined by the representer theorem~\cite{rth01}. Section~\ref{dskernel} presents our proposed distributional semantic kernel  between unstructured text description, which is applicable to  our kernel formulation and can be useful for other applications as well. Section~\ref{experiments} presents our experiments  on Flower Dataset~\cite{Flower08} and Caltech-UCSD dataset~\cite{CU20010} for both the linear and the kernel classifier predictions.


\section{Related Work}
\label{relwork}

We focus our related work discussion on three related lines of research: ``zero/few-shot learning'', ``visual knowledge transfer'', and ``Language and Vision''.


\textbf{Zero/Few-Shot Learning: }
Motivated by the practical need to learn visual classifiers of rare categories, researchers have explored approaches for learning from a single image (one-shot learning~\cite{Miller2000,fe2003bayesian,Fink04,BartU05}) or even from no images (zero-shot learning).
One way of recognizing object instances from previously unseen test categories
(the zero-shot learning problem) is by leveraging knowledge about common attributes and shared parts.
Typically an intermediate semantic layer is introduced to enable sharing knowledge between classes and facilitate describing knowledge about novel unseen classes, \eg~\cite{Palatucci09}.
 For instance, given adequately labeled training data, one can learn classifiers for the attributes occurring in the training object categories. These classifiers can then be used to recognize the same attributes in object instances from the novel test categories. Recognition can then proceed on the basis of these learned attributes~\cite{Lampert09, Farhadi09}. Such attribute-based ``knowledge transfer'' approaches use an intermediate visual attribute representation to enable describing unseen object categories. 
 
 Typically attributes~\cite{Lampert09,Farhadi09} are manually defined by humans to describe shape, color, surface material, \eg, furry, striped, \etc    ~Therefore, an unseen category has to be specified in terms of the used vocabulary of attributes.  Rohrbach \etal~\cite{rohrbach10eccv} investigated extracting useful attributes from large text corpora. In~\cite{ParikhG11},  an approach was introduced for interactively defining a vocabulary of attributes that are both human understandable and visually discriminative. Huang \etal~\cite{Huang_2015_CVPR} relaxed the attribute independence assumption by modeling correlation between attributes to achieve better zero shot performance, as opposed to prior models.

 Similar to the setting of zero-shot learning, we use classes with training data (seen classes) to predict classifiers for classes with no training data (unseen classes).  
 In contrast to attributes based method (e.g., ~\cite{Lampert09,Farhadi09}), in our work we do not use any explicit attributes. The description of a new category is purely textual and the process is completely automatic without human annotation beyond the class labels.


\textbf{Visual Knowledge Transfer: }Our work can be seen in the context of knowledge sharing and inductive transfer. 
In general, knowledge transfer aims at enhancing recognition by exploiting shared knowledge between classes. Most existing research focused on knowledge sharing within the visual domain only, \eg~\cite{griffin2008learning}; or exporting semantic knowledge at the level of category similarities and hierarchies, \eg~\cite{fergus2010semantic,Salakhutdinov11}.  We go beyond the state-of-the-art to explore cross-domain knowledge sharing and transfer. We explore how knowledge from the visual and textual domains can be used to learn across-domain correlation, which facilitates prediction of visual classifiers from textual description.




\textbf{Language and Vision:} The relation between linguistic semantic representations and visual recognition has been explored. For example in~\cite{deng2010does}, it was shown that there is a strong correlation between semantic similarity between classes, based on WordNet, and confusion between classes. Linguistic semantics in terms of nouns from WordNet \cite{wordNet95} have been used in collecting large-scale image datasets such as ImageNet\cite{imNet09} and Tiny Images~\cite{tinyimages}. It was also shown that hierarchies based on WordNet are useful in learning visual classifiers, \eg~\cite{Salakhutdinov11}.


One of the earliest work on learning from images and text corpora is the work of Barnard \etal~\cite{barnard2001clustering}, which showed that learning  a joint distribution of words and visual elements facilitates clustering the images in a semantic way, generating illustrative images from a caption, and generating annotations for novel images.
There has been an increasing recent interest in the intersection between computer vision and natural language processing with researches that focus on generating textual description of images and videos, \eg~\cite{farhadi2010every,kulkarni2011baby,yang2011corpus,krishnamoorthy2013generating}. This includes generating sentences about objects, actions, attributes, spatial relation between objects, contextual information in the images, scene information, \etc
        ~Based on the success of sequence to sequence training of neural nets 
in  machine translation  
(e.g.,~\cite{cho2014learning}), impressive works has been recently proposed for image captioning (e.g., ~\cite{karpathy2014deep,vinyals2015show,xu2015show,mao2015deep}). In contrast, our work is different in two fundamental ways. In terms of the goal, we do not target generating textual description from images, instead we target  predicting classifiers from text, in a zero-shot setting. In terms of the learning setting, the textual descriptions that we use is at the level of the category  and do not come in the form of image-caption pairs, as in typical datasets used for text generation from images, \eg~\cite{ordonez2011im2text}.   

There are several recent works that studies unannotated text with images\ignore{ \cite{NIPS13DeViSE,NIPS13CMT,Hoseini13}}.  In \cite{NIPS13DeViSE,NIPS13CMT},  word  embedding language models (\eg ~\cite{mikolov2013distributed}) were adopted to represent class names as vectors, which require training using a  big text-corpus. Their goal is to embed images into the language space then perform classification. In  ~\cite{elhoseiny2016zero}, a similar yet multimodal approach was adopted for Multimedia Event Detection in videos instead of object classification. There are several differences between these works and our method. First, one limitation of the adopted language model is that it produces only one vector per word, which causes problems when a word has multiple meanings. Second, these methods assumes that each class is represented by one or few-words and hence  can not represent a class text description that typically contains multiple paragraphs in our setting. Third,  our goal is different which is to map the text description to an explicit classifier in the visual domain, \ie the opposite direction of their goal. Fourth, these models do not support non-linear classification, supported by the kernelized version proposed in this work\ignore{, which is supported by our method}. Finally, we focus on fine-grained recognition, which is a very  challenging task\ignore{; see figure~\ref{fig:problem}}.  

\section{Problem Definition}
\label{obvformulation}

Fig~\ref{F:prob_def} illustrates the learning setting. 
The information in our problem comes from two different domains: the visual domain  and the textual domain, denoted by $\mathcal{V}$ and $\mathcal{T}$, respectively. Similar to traditional visual learning problems, we are given training data in the form $V=\{({x}_i , l_i)\}_{N}$, where $x_i$ is an image and $l_i \in \{1\cdots N_{sc}\}$ is its class label. We denote the number of classes available at training as $N_{sc}$, where $sc$ indicates ``seen classes''. As typically done in visual classification setting, we can learn $N_{sc}$ binary one-vs-all classifiers, one for each of these classes.  

Our goal is to be able to predict a classifier for a new category based only on the learned classes and a textual description(s) of that category. In order to achieve that, the learning process has to also include textual description of the seen classes (as shown in Fig ~\ref{F:prob_def} ). Depending on the domain we might find a few, a couple, or as little as one textual description to each class. We denote the textual training data for class $j$ by $\{t_i\in \mathcal{T} \}^j$. 
In this paper we assume we are dealing with the extreme case of having only one textual description available per class, which makes the problem even more challenging. For simplicity, the text description of class $j$ is denoted by $t_j$. However, the formulation we propose in this paper directly applies to the case of multiple textual descriptions per class.

In this paper, we discuss the task of predicting  visual classifier ${\Phi}(t_*)$ from an unseen text description $t_*$  in  linear form or RKHS kernalized form, defined as follows

\subsection{Linear Classifier}
\label{sec_lin_pdef}
Let us consider a typical \ignore{binary }linear classifier in the feature space in the form
\[
    f_j(\mathbf{x}) = \mathbf{c}_j^\textsf{T} \cdot \mathbf{x}
\] 
where $\mathbf{x}$ (bold) is the visual feature vector of an image $x$ (not bold) amended with 1   and $\mathbf{c}_j \in \mathbb{R}^{d_v}$ is the linear classifier parameters for class $j$. Given a test image, its class is determined by
\begin{equation}
\small
l^* = \arg \max_j f_j(\mathbf{x})
\label{eq:mclass}
\end{equation}
  
 Similar to the visual domain, the raw textual descriptions have to go through a feature extraction process\ignore{, which will be described in Sec~\ref{experiments}}. Let us denote the linear extracted textual feature by 
$T=\{\mathbf{t}_j \in \mathbb{R}^{d_t}\}_{j=1\cdots N_{sc}}$, where $\mathbf{t}_j$ is the features of text description $t_j$ (not bold).  Given a textual description ${t}_*$  of a new unseen category $\mathcal{U}$  with linear feature vector representation $\mathbf{t}_*$, the problem can now be defined as predicting a one-vs-all linear classifier parameters ${\Phi}{(t_*)} = c(\mathbf{t}_*) \in \mathbb{R}^{d_v}$,  such that it can be directly used to classify any test image $\mathbf{x}$ as (also see Table~\ref{tbl_class_pred})
\begin{eqnarray}
      c(\mathbf{t}_*)^\textsf{T} \cdot \mathbf{x} >  0 & \text{if} \;  \mathbf{x} \; \text{belongs to} \;\mathcal{U} \nonumber \\
      c(\mathbf{t}_*)^\textsf{T} \cdot \mathbf{x} <  0 &  \text{otherwise} \label{E:PredictedClass}
\end{eqnarray}


\subsection{Kernel Classifier}
\label{sec_kernel_pdef}
For kernel classifiers, we assume that each of the domains is equipped with a kernel function corresponding to a \textit{reproducing kernel Hilbert space} (RKHS). Let us denote the kernel for $\mathcal{V}$ by $k(\cdot,\cdot)$, and the kernel for $\mathcal{T}$ by $g(\cdot,\cdot)$. 
\ignore{Since, we are studying explicit kernel-classifier prediction from privileged information, we first present an overview on multi-class classification on kernel space.  One} 
\normalsize 

According to the generalized representer theorem~\cite{rth01},  a minimizer of a regularized empirical risk function over an RKHS could be represented as a linear combination of kernels, evaluated on the training set. Adopting the representer theorem on classification risk function, we define a kernel-classifier of a visual class $j$ as follows

\begin{table}[t!]
\centering
\caption{Classifier Prediction Functions (Linear and Kernel)}
\label{tbl_class_pred}
\vspace{-1mm}
\begin{tabular}{|c|c|}
\hline 
\textbf{Linear Prediction }& \textbf{Kernel Prediction }\\ 
\hline 
${\Phi}(t_*) = c(\mathbf{t}_*)$& ${\Phi}(t_*) = \boldsymbol{\beta}({t}_{*})$ \\ 
\hline 
\end{tabular} 
\vspace{-4mm}
\end{table}

\begin{equation}
\small
\begin{split}
f_j({x})=&   \sum_{i=1}^{N} \beta_j^i k({x}, {x}_i) + b  = \sum_{i=1}^{N} \beta_j^i \varphi({x_i})^\textsf{T} \varphi( {x}) + b \\
f_j({x})=&  {\boldsymbol{\beta}_j}^\textsf{T} \cdot  \textbf{k}({x}) =  \mathbf{c}_j^\textsf{T} \cdot [\varphi(x); 1] , \mathbf{c}_j =  [\sum_{i=1}^{N} \beta_j^i \varphi({x_i}); b]\\
\end{split}
\label{eq_fkernel}
\end{equation}
where ${x} \in \mathcal{V}$ is the test image, ${x}_i$ is the $i^{th}$ image in the training data $V$,  $\textbf{k}({x})= [k({x}, {x}_1), \cdots, k({x}, {x}_N), 1]^\textsf{T},$  $\boldsymbol{\beta}_j = [\beta_j^1 \cdots \beta_j^N, b]^\textsf{T} $. Having learned $f_j({x}^*)$ for each class $j$ (for example using SVM classifier), the class label of the test image ${x}$ can be predicted by  Eq.~\ref{eq:mclass}, similar to the linear case. Eq.~\ref{eq_fkernel} also shows how $\boldsymbol{\beta}_j$ is related to $\mathbf{c}_j$ in the linear classifier, where $k(x, x')= \varphi(x)^\mathsf{T} \cdot \varphi(x')$ and $\varphi(\cdot)$ is a feature map  that does not have to be explicitly defined given the  definition of $k(\cdot, \cdot)$ on $\cal{V}$. Hence, our goal in the kernel classifier prediction  is to predict $\boldsymbol{\beta}(t_*)$ instead of $\mathbf{c}(t_*)$ since it is sufficient to define  ${f}_{t_*}(x)$ for a text description $t_*$ of an unseen class given $\mathbf{k}(x)$


\ignore{Under our setting, i}It is clear that $f_j({x})$ could be learned for  all classes with training data $j \in  {1} \cdots {N_{sc}}$, since there are examples for the seen classes; we denote the kernel-classifier parameters of the seen classes as $\mathcal{B}_{sc} =  \{  \boldsymbol{\beta}_j \}_{N_{sc}}, \forall j$. However, it is not obvious how to predict $f_{{t}_*}({x})$ for an unseen class given its text description ${t}_*$. Similar to the linear classifier prediction, our main notion is to use the text description ${t}_{*}$, associated with unseen class, and the training data to directly predict the unseen kernel-classifier parameters. In other words, the kernel classifier parameters of the unseen class is a function of  its text description ${t}_*$ , the image training data $V$ and the text training data $\{t_j\}, j\in 1 \cdots N_{sc}$; \ie \small
\[   f_{{t}_*}({x}) = \boldsymbol{\beta}({t}_*)^\textsf{T} \cdot \textbf{k}({x}), \] \normalsize
 $ f_{{t}_*}({x})$ could be used to classify new points that	 belong to an  unseen class as follows: 1) one-vs-all setting  $f_{{t}_*}({x})  \gtrless  0$  \ignore{if ${x}^*\,$  belongs to unseen category $z^*$, $\boldsymbol{\beta}(\mathbf{t}_*),)^\textsf{T} \cdot \textbf{k}(x^*)< 0$  otherwise}; or 2) in a Multi-class prediction as in Eq~\ref{eq:mclass}. In this case, ${\Phi}(t_*)=\boldsymbol{\beta}({t}_{*})$; see Table~\ref{tbl_class_pred}. In contrast to the linear classifier prediction, there is no need to explicitly represent an image $x$ or a text description $t$ by features, which are denoted by the bold symbols in the previous section. Rather, only  $k(\cdot,\cdot)$  and $g(\cdot,\cdot)$  must   be defined which leads to more general classifiers.

\section{Relation to Regression and Knowledge Transfer Models}
\label{sec_reganKT}
We introduce two possible frameworks for this problem and discuss potential limitations for them. In this background section,  we focus on predicting linear classifiers for simplicity, which motivates the evaluated linear classifier formulations that follow in Sec~\ref{formulation}. 

\subsection{Regression Models}
A straightforward way to solve this problem is to pose it as a regression problem where the goal is to use the textual data and the learned classifiers, $\{(\mathbf{t}_j,\mathbf{c}_j) \}_{j=1\cdots N_{sc}}$ to learn a regression function from the textual feature domain to the visual classifier domain, \ie, a function $c(\cdot) : \mathbb{R}^{d_t} \rightarrow \mathbb{R}^{d_v} $.  The question is which regression model would be suitable for this problem? and would posing the problem in this way lead to reasonable results?

A typical regression model, such as ridge regression~\cite{ridgeReg70} or Gaussian Process (GP) Regression~\cite{Rasmussen:2005}, learns the regressor to each dimension of the output domain (the parameters of a linear classifier) separately, \ie,  a set of functions $c^i(\cdot) : \mathbb{R}^{d_t} \rightarrow \mathbb{R} $. Clearly this will not capture the correlation between the visual classifier dimensions. Instead, a structured prediction regressor would be more suitable since it would learn the correlation between the input and output domain. However, even a structured prediction model will only learn the correlation between the textual and visual domain through the information available in the input-output pairs  $(\mathbf{t}_j,\mathbf{c}_j)$.  Here the visual domain information is encapsulated in the pre-learned classifiers and prediction does not have access to the original data in the visual domain. Instead, we need to directly learn the correlation between the visual and textual domain and use that for prediction.

Another fundamental problem  that a regressor would face, is the sparsity of the data; the data points are the textual description-classifier pairs, and typically the number of classes can be very small compared to the dimension of the classifier space (\ie $N_{sc} \ll d_v$). In a setting like that, any regression model is bound to suffer from an under fitting problem. This can be best explained in terms of GP regression, where the predictive variance increases in the regions of the input space where there are no data points. This will result in  poor prediction of classifiers at such  regions. 

\subsection{Knowledge Transfer Models}
An alternative formulation is to pose the problem as domain adaptation from the textual to the visual domain. In the computer vision context, domain adaptation work has focused on transferring categories learned from a source domain,  with a given distribution of images, to a target domain with a different distribution, \eg, images or videos from different sources~\cite{yang07,saenko10,da11,duan12}. 
What we need is an approach that learns the correlation between the textual domain features and the visual domain features, and uses that correlation to predict new visual classifier given textual features. 

In particular, in~\cite{da11} an approach for learning cross domain transformation was introduced. In that work a regularized asymmetric transformation between points in two domains were learned. The approach was applied to transfer learned categories between different data distributions, both in the visual domain. A particular attractive characteristic of~\cite{da11}, over other domain adaptation models, is that the source and target domains do not have to share the same feature spaces or the same dimensionality. 

While a totally different setting is studied in ~\cite{da11}, it inspired us to formulate the zero-shot learning problem as a domain transfer problem. This can be achieved by learning a linear transfer function $\mathbf{W}$ between $\mathcal{T}$ and $\mathcal{V}$.  The transformation matrix $\mathbf{W}$ can be learned  by optimizing, with a suitable regularizer, over constraints of the form $\mathbf{t}^\textsf{T}\mathbf{W}\mathbf{x} \geq l $ if $\mathbf{t} \in \mathcal{T}$  and $\mathbf{x}\in \mathcal{V}$  belong to the same class, and $\mathbf{t}^\textsf{T}\mathbf{W}\mathbf{x} \leq u $  otherwise. Here $l$ and $u$ are model parameters. This transfer function acts as a compatibility function between the textual features and visual features, which gives high values if they are from the same class and a low value if they are from different classes. 


It is not hard to see that this transfer function can act as a classifier. Given a textual feature $\mathbf{t}^*$ and a test image, represented by $\mathbf{x}$, a classification decision can be obtained by $\mathbf{t}_*^\textsf{T}\mathbf{W}\mathbf{x} \gtrless b$ where $b$ is a decision boundary which can be set to $(l+u)/2$. Hence, our desired predicted classifier in Eq~\ref{E:PredictedClass} can be obtained as $c(\mathbf{t}_*) = \mathbf{t}_*^\textsf{T}\mathbf{W} $ (note that the features vectors are amended with ones).  However, since learning  $\mathbf{W}$ was done over seen classes only, it is not clear how the predicted classifier $c(\mathbf{t}_*)$ will behave for unseen classes. There is no guarantee that such a classifier will put all the seen data on one side and the new unseen class on the other side of that hyperplane.


%


\section{Formulations for Predicting a linear   classifier form ( ${\Phi (t_*)} = c(\mathbf{t}_*)$)}
\label{formulation}
The proposed formulations in this section aims at predicting a linear hyperplane parameter $\mathbf{c}$ of a one-vs-all classifier for a new unseen class given a textual description, encoded as a feature vector $\mathbf{t_*}$ and the knowledge learned at the training phase from seen classes\footnote{The notations follow from Subsection~\ref{sec_lin_pdef}}. We start by  defining the learning components that are used by the formulations described in this section:

\begin{description}
\item [Classifiers:]$\,\,\,\,\,\,\,\,$ 

a set of linear one-vs-all classifiers $\{\mathbf{c}_j\}$ are learned, one for each seen class.
\item [Probabilistic Regressor:] $\,\,\,\,\,\,\,\,\,\,\,\,$

Given $\{(\mathbf{t}_j,\mathbf{c}_j)\}$ a regressor is learned that can be used to give a prior estimate for $p_{reg}(\mathbf{c} | \mathbf{t})$ (Details in Sec~\ref{S:Reg}).
\item [Domain Transfer:]$\,\,\,\,\,\,\,\,\,\,\,\,$ 

Given $T$ and $V$, a domain transfer function encoded in the matrix $\mathbf{W}$, is learned  which captures the correlation between the textual and visual domains (Details in Sec~\ref{S:DA}).
\end{description}

Each of the following subsections show a different approach to predict a linear classifier from $t_*$ as $\Phi(t_*) = \mathbf{c}(\textbf{t}_*)$; see Sec~\ref{sec_lin_pdef}. The final approach (E), which achieves the best performance, combines regression, domain transfer, and additional constraints. We compare between these alternative formulations (A to E) in our experiments. Hyper-parameter selection is detailed in the supplementary materials for all the approaches. 

\begin{figure*}[t]
\centering
\includegraphics[width=0.85\linewidth,height=.38\linewidth]{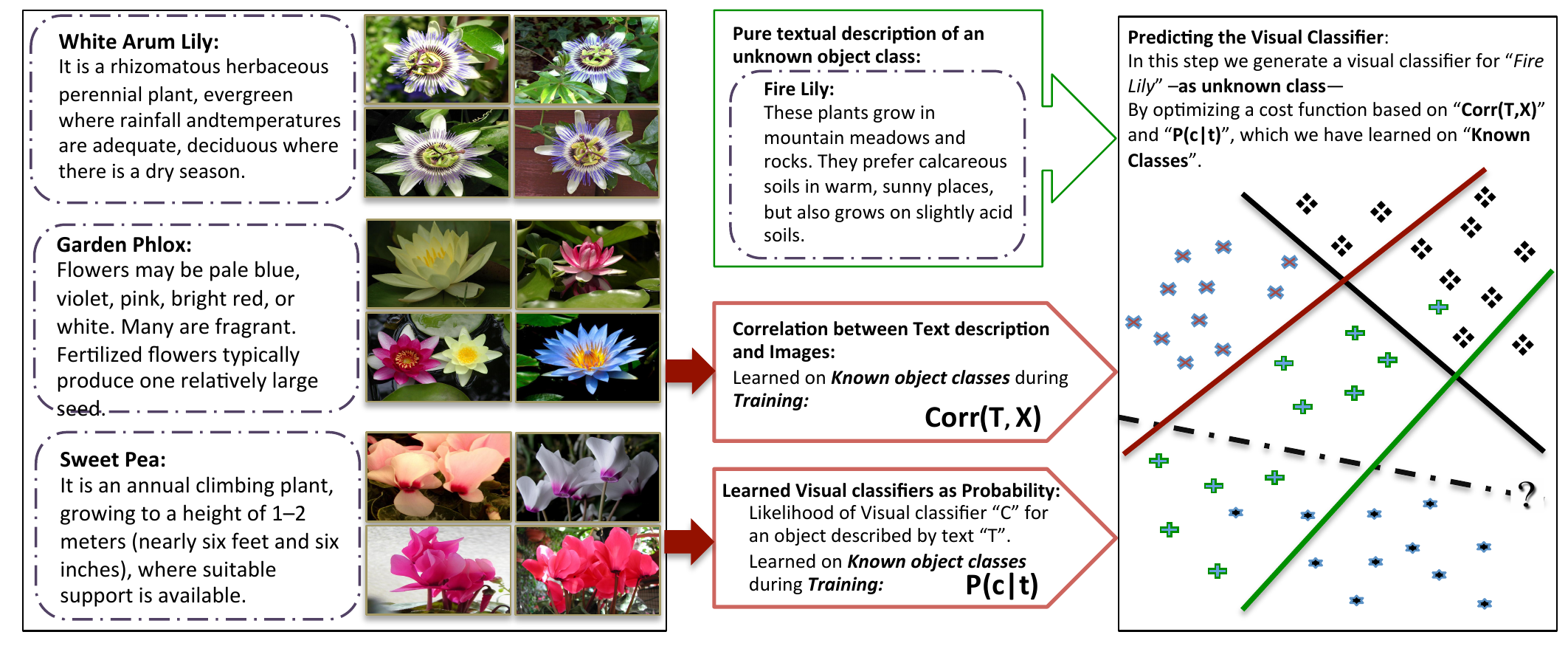}
\vspace{-10pt}
\caption{Illustration of the Proposed Linear Prediction Framework (Constrained Regression and Domain Transfer) for the task Zero-shot learning from textual description (Linear Formulation (E))}
\label{F:prob_sol}
\vspace{-12pt}
\end{figure*}

\subsection{Probabilistic Regressor}
\label{S:Reg}
There are different regressors that can be used, however we need a regressor that provide a probabilistic estimate $p_{reg}(\mathbf{c} | \mathbf(t))$. For the reasons explained in Sec~\ref{obvformulation}, we also need a structure prediction approach that is able to predict all the dimensions of the classifiers together. For these reasons, we use the Twin Gaussian Process (TGP)~\cite{Bo:2010}. 
TGP encodes the relations between both the inputs and structured outputs using Gaussian Process priors. This is achieved by minimizing the Kullback-Leibler divergence between the marginal GP of the outputs (i.e. classifiers in our case) and observations (i.e. textual features). The estimated regressor output ($\tilde{c}(\mathbf{t}_*)$) in TGP is given by the solution of the following non-linear optimization problem~\cite{Bo:2010} \footnote{notice we are using $\mathbf{\tilde{c}}$ to denote the output of the regressor, while using $\mathbf{\hat{c}}$ to denote the output of the final optimization problem in Eq~\ref{eq:form}}.
\begin{equation}                                               
\centering
\begin{split}
{\Phi (t_*)} = \tilde{c}(\mathbf{t}_*) = & \underset{\mathbf{c}}{\operatorname{argmin}}[  K_C(\mathbf{c},\mathbf{c})  -2 k_c(\mathbf{c})^\textsf{T} \mathbf{u} - \eta  \log ( \\ & K_C(\mathbf{c},\mathbf{c} -k_c(\mathbf{c})^\textsf{T} (\mathbf{K}_C+ \lambda_c \mathbf{I})^{-1} k_c(c) ) ]
\end{split}
\label{eq:tgp}
\end{equation}
where $\mathbf{u} = (\mathbf{K}_\textsf{T} + \lambda_t  \mathbf{I})^{-1} k_t(\mathbf{t}_*)$, $\eta  = K_T(\mathbf{t}_*,\mathbf{t}_*) -k(\mathbf{t}_*)^\textsf{T}  \mathbf{u} $,  and  $K_T(\mathbf{t}_l,\mathbf{t}_m) $ and $K_C(\mathbf{c}_l,\mathbf{c}_m)$ are Gaussian kernel for input feature $\mathbf{t}$ and output vector $\mathbf{c}$, respectively.  $k_c(\mathbf{c}) = [K_C(\mathbf{c},\mathbf{c}_1), \cdots, K_C($ $\mathbf{c},\mathbf{c}_{N_{sc}})]^\textsf{T}$. $k_t(\mathbf{t}_*) = [K_T(\mathbf{t}_*,\mathbf{t}_1), \cdots, K_T(\mathbf{t}_*,\mathbf{t}_{N_{sc})}]^\textsf{T}$.  $\lambda_t$ and $\lambda_c$ are regularization parameters to avoid overfitting. This optimization problem can be solved using a second order, BFGS quasi-Newton optimizer with cubic polynomial line search for optimal step size selection~\cite{Bo:2010}. In this case, the classifier dimensions are predicted jointly. Hence, $p_{reg}(\mathbf{c}|\mathbf{t}_*)$ is defined as a normal distribution.
\begin{equation}
\label{eq:ptgp}
p_{reg}(\mathbf{c}|\mathbf{t}_*) =  \mathcal{N} (\mu_c = \tilde{c}(\mathbf{t}_*),\Sigma_c = \mathbf{I})
\end{equation}
The reason that $\Sigma_c = \mathbf{I}$ is that TGP does not provide predictive variance, unlike Gaussian Process Regression. However, it has the advantage of handling the dependency between the dimensions of the classifiers $\mathbf{c}$ given the textual features $\mathbf{t}$. 
 
\subsection{Constrained Probabilistic Regressor}
\label{S:Reg_con}
We also investigated formulations that use regression to predict an initial hyperplane  $\tilde{c}(\mathbf{t}_*)$ as described in section ~\ref{S:Reg}, which is then optimized to put all seen data in one side, \ie
\begin{equation*}
\begin{split}
 {\Phi (t_*)} = \small \hat{c}(\mathbf{t}_*) =   \underset{\mathbf{c},\zeta_i}{\operatorname{argmin }}[  \mathbf{c}^\textsf{T} \mathbf{c}  + \alpha \, \psi( \mathbf{c},\tilde{c}(\mathbf{t}_*)) + C   \sum_{i=1}^N{\zeta_i} ] \\ \; s.t.:  -\mathbf{c}^\textsf{T} {\mathbf{x}}_{i}  \geq \zeta_i , \; \zeta_i \geq 0 , i=1,\cdots,N 
\end{split}
\end{equation*}

where $\psi(\cdot,\cdot)$ is a similarity function between hyperplanes, e.g., a dot product used in this work,  $\alpha$ is its constant weight, and $C$ is the weight to the soft constraints of existing images as negative examples (inspired by linear SVM formulation)\ignore{, or other functions incorporating the predictive variance}. We call this class of methods {\em constrained GPR/TGP}, since $\tilde{c}(\mathbf{t}_*)$ is initially predicted through GPR or TGP.

\subsection{Domain Transfer (DT)}
\label{S:DA}
To learn the domain transfer function $\mathbf{W}$ we adapted the approach in~\cite{da11} as follows. Let $\mathbf{T}$ be the textual feature data matrix and $\mathbf{X}$ be the visual feature data matrix where each feature vector is amended with a 1. Notice that amending the feature vectors with a 1 is essential in our formulation since we need $\mathbf{t}^\textsf{T} \mathbf{W}$ to act as a classifier. We need to solve the following optimization problem 
\begin{equation}
  \min_{\mathbf{W}}  r(\mathbf{W}) + \lambda \sum_i c_i(\mathbf{T} \mathbf{W} \mathbf{X}^\textsf{T})
  \label{Eq:DA1}
\end{equation}
where $c_i$'s are loss functions over the constraints and $r(\cdot)$ is a matrix regularizer.  It was shown in~\cite{da11}, under condition on the regularizer, that the optimal $\mathbf{W}$  is in the form of 
 $ \mathbf{W}^* = \mathbf{T} \mathbf{K}_{T}^{-\frac{1}{2}} \mathbf{L}^*  \mathbf{K}_{X}^{-\frac{1}{2}} \mathbf{X}^\textsf{T}$, where  $\mathbf{K}_{T}  = \mathbf{T} \mathbf{T}^\textsf{T}$,  $\mathbf{K}_{X}  = \mathbf{X} \mathbf{X}^\textsf{T}$.   $\mathbf{L}^*$ is computed by minimizing the following minimization problem
\begin{equation}
 \underset{\mathbf{L}}{\operatorname{min       }}[   r(\mathbf{L})+ \lambda \sum_p c_p(\mathbf{K}_{T}^{\frac{1}{2}} \mathbf{L} \mathbf{K}_{X}^\frac{1}{2}  ) ],
\end{equation}
where $c_p(\mathbf{K}_{T}^{\frac{1}{2}} \mathbf{L} \mathbf{K}_{X}^\frac{1}{2}  ) = (max(0, (l-e_i \mathbf{K}_{T}^{\frac{1}{2}} \mathbf{L} \mathbf{K}_{X}^\frac{1}{2} e_j) ))^2$ for same class pairs of index $i$,$j$, or $ =(max(0, (e_i \mathbf{K}_{T}^{\frac{1}{2}} \mathbf{L} \mathbf{K}_{X}^\frac{1}{2} e_j -u) ))^ 2$ otherwise, where $e_k$ is a one-hot vector of zeros except a one at the $k^{th}$ element, and $u>l$. In our work, we used $l =2$, $u=-2$ (note any appropriate $l$ and $u$ can work). We used a Frobenius norm regularizer.  This energy is minimized using a second order BFGS quasi-Newton optimizer. Once $L$ is computed $\textbf{W}^*$ is computed using the transformation above. Finally ${\Phi (t_*)}  = c(\mathbf{t}_*) = \mathbf{t}_*^\textsf{T}\mathbf{W}$, simplifying  $\textbf{W}^*$ as $\textbf{W}$.

\subsection{Constrained-DT}
We also investigated constrained-DT formulations that learns a transfer matrix $\mathbf{W}$ and  enforce $\mathbf{t}_j^\textsf{T}\mathbf{W}$ to be close to the classifiers learned on seen data, $\{\mathbf{c}_j \}$ ,\ie
\begin{equation*}
\small \min_{\mathbf{W}}  r(\mathbf{W}) + \lambda_1 \sum_i c_i(\mathbf{T} \mathbf{W} \mathbf{X}^\textsf{T}) +\lambda_2 \sum_j{\|\mathbf{c}_j - \mathbf{t}^\textsf{T}_j \mathbf{W} \|^2}
\end{equation*}
A classifier can be then obtained by ${\Phi (t_*)}  =c(\mathbf{t}_*)= \mathbf{t}_*^\textsf{T}\mathbf{W}$.

\subsection{Constrained  Regression and Domain Transfer for classifier prediction}
 Fig~\ref{F:prob_sol} illustrates our final  framework which combines regression (formulation A (using TGP)) and domain transfer (formulation C) with additional constraints. This formulation combines the three learning components described in the beginning of this section. 
Each of these components contains partial knowledge about the problem. The question is how to combine such knowledge to predict a new classifier given a textual description. The new classifier has to be consistent with the seen classes. 
The new classifier has to put all the seen instances at one side of the hyperplane, and has to be consistent with the learned domain transfer function. This leads to the following constrained  optimization problem 
\begin{equation}
\begin{split}
 {\Phi (t_*)} = \hat{c}(\mathbf{t}_*) =  &   \underset{\mathbf{c},\zeta_i}{\operatorname{argmin }}\big[    \mathbf{c}^\textsf{T} \mathbf{c} - \alpha {\mathbf{t}_*}^\textsf{T} \mathbf{W} \mathbf{c}  - \gamma  \ln( p_{reg}(\mathbf{c}|\mathbf{t}_*))  \\
& + C   \sum{\zeta_i} \big]\\
&s.t.:  -(\mathbf{c}^\textsf{T} {\mathbf{x}}_{i} ) \geq \zeta_i ,  \,\, \zeta_i \geq 0 ,\; \; i = 1 \cdots N \\
&	   \,\, {\mathbf{t}_*}^\textsf{T} \mathbf{W} \mathbf{c} \geq l     \\
& \alpha , \gamma, C, l \,  \text{: hyperparameters}\\ 
\end{split}
\label{eq:form}
\end{equation}
The first term is a regularizer over the classifier $\mathbf{c}$.  The second term enforces that the predicted classifier has high correlation with $\mathbf{t}_*^\textsf{T} \mathbf{W}$; $\mathbf{W}$ is learnt by Eq~\ref{Eq:DA1}. The third term favors a classifier that has high probability given the prediction of the regressor. The constraints $ -\mathbf{c}^\textsf{T} {\mathbf{x}}_{i} \geq \zeta_i $  enforce all the seen data instances to be at the negative side of the predicted classifier hyperplane with some missclassification allowed through the  slack variables $ \zeta_i$. The constraint $ {\mathbf{t}_*}^\textsf{T} \mathbf{W} \mathbf{c} \geq l$ enforces that the correlation between the predicted classifier and ${\mathbf{t}_*}^\textsf{T} \mathbf{W}$ is no less than $l$, this is to enforce a minimum correlation between the text and visual features.




\textbf{Solving for $\hat{c}$ as a quadratic program: }
According to  the definition of $p_{reg}(\mathbf{c}|\mathbf{t}_*)$ for  TGP,  $\ln p(\mathbf{c}|\mathbf{t}_*)$ is a quadratic term in $c$ in the form  
\begin{equation}
\begin{split}
-\ln p(\mathbf{c}|\mathbf{t}_*) \propto ( \mathbf{c} - \tilde{c}(\mathbf{t}_*))^\textsf{T} (\mathbf{c} - \tilde{c}(\mathbf{t}_*)) \\= \mathbf{c}^\textsf{T} \mathbf{c} -2 \mathbf{c}^\textsf{T} \tilde{c}(\mathbf{t}_*) +  \tilde{c}(\mathbf{t}_*)^\textsf{T} \tilde{c}(\mathbf{t}_*) 
\end{split}
\label{eq:lnptgp}
\end{equation}
We reduce $-\ln p(\mathbf{c}|\mathbf{t}_*)$ to $-2 \mathbf{c}^\textsf{T} \tilde{c}(\mathbf{t}_*))$, since 1) $\tilde{c}(\mathbf{t}_*)^\textsf{T} \tilde{c}(\mathbf{t}_*)$ is a constant (\ie does not affect the optimization), 2) $\mathbf{c}^\textsf{T} \mathbf{c}$ is already included as regularizer in equation ~\ref{eq:form}.  In our setting, the dot product  is a better similarity measure between two hyperplanes. Hence,  $-2 \mathbf{c}^\textsf{T} \tilde{c}(\mathbf{t}_*)$ is minimized.
Given $-\ln p(\mathbf{c}|\mathbf{t}_*)$ from the TGP and  $\mathbf{W}$, Eq~\ref{eq:form} reduces to a quadratic program on $\mathbf{c}$ with linear constraints. We tried different quadratic solvers, however the IBM CPLEX solver \footnote{http://www-01.ibm.com/software/integration/optimization/cplex-optimizer} gives the best performance in speed and optimization for our problem.

\section{Formulations for Predicting a kernel   classifier form ( ${\Phi}(t_*) = \boldsymbol{\beta}({t}_{*})$ )}
\label{sec:app}

Prediction of ${\Phi (t_*)}  = \boldsymbol{\beta}(t_*)$ (Sec.~\ref{sec_kernel_pdef}), is decomposed into training (domain transfer) and prediction phases, detailed as follows
\subsection{Kernelized Domain Transfer}
\label{ss:tr}
During training, we firstly learn $\mathcal{B}_{sc} = \{ \boldsymbol{\beta}_j \}, j=1\to N_sc$ as SVM-kernel classifiers based on  the training data and defined by $k(\cdot, \cdot)$ visual kernel\ignore{, see Sec~\ref{sec:pdef}}. Then, we learn a kernel domain transfer function to transfer the text description information ${t_*} \in \mathcal{T}$ to kernel-classifier parameters $\boldsymbol{\beta} \in \mathbb{R}^{N+1}$ in $\mathcal{V}$ domain. We call this domain transfer function $\boldsymbol{\beta}_{DA}(t_*)$, which has the form of ${\mathbf{\Psi}}^\textsf{T} {\textbf{g}(t_*)}$, where $\textbf{g}(t_*)  = [g({t_*}, {t}_1) \cdots g({t_*}, {t}_{N_{sc}})]^\textsf{T}$\ignore{, $g({t}, {t}')\,$\normalsize is a kernel function that measures the similarity between ${t}$ and ${t}'$ on  domain $\mathcal{E}$}; $\mathbf{\Psi}$ is an $N_{sc}  \times {N+1}$ matrix, which transforms ${t}$ to  kernel classifier parameters for the class that ${t_*}$ represents.

\ignore{Inspired by the domain transfer proposed by~\cite{da11}, w}We aim to learn  $\mathbf{\Psi}$ from $V$ and $\{t_j\}, j=1 \cdots N_{sc}$, such that $\textbf{g}(t)^\textsf{T} \mathbf{\Psi} \textbf{k}(x) > l$ if ${t}$ and  $x\,$ correspond to the same class, $\textbf{g}(t)^\textsf{T} {\mathbf{\Psi}} \textbf{k}(x) < u\,$ otherwise. Here $l\,$ controls similarity lower-bound if $t$ and $x$ correspond to  same class, and $u$ controls similarity upper-bound if $t$ and $x$ belong to different classes. In our setting, the term   ${\mathbf{\Psi}}^\textsf{T} {\textbf{g}(t_j)}$ should act as a classifier parameter for class $j$ in the training data. Therefore,  we introduce  penalization constraints to our minimization function  if  ${\mathbf{\Psi}}^\textsf{T}\,{\textbf{g}(t_j)}$ is distant from $\boldsymbol{\beta}_j \in \mathcal{B}_{sc}$, where ${t}_i$ corresponds to the class that $\boldsymbol{\beta}_i$ classifies.\ignore{Hence,  in order to learn \small${\textbf{T}}\,$\normalsize, we solve the following objective function  Inspired by domain adaptation \footnote{A totally different problem/setting but the optimization methods inspired our solution} optimization methods (\eg \cite{da11}),  we model our solution using the following objective functionInspired by domain adaptation optimization methods (\eg \cite{da11}) }\ignore{ \footnote{A totally different problem/setting but the optimization methods inspired our solution},  in order to learn ${\textbf{T}}$,,} we model the kernel domain transfer function as follows \ignore{by the following objective function}
\small
\begin{equation}
\begin{split}
{\mathbf{\Psi}^*}= 
 \arg \min_{{\mathbf{\Psi}}}  L({\mathbf{\Psi}}) = [&\frac{1}{2} r({\mathbf{\Psi}}) + \lambda_1 \sum_k c_k(\mathbf{G}\, {\mathbf{\Psi}}\, \mathbf{K}) + \\ & \lambda_2 \sum_{i=1}^{N_{sc}}{\|\boldsymbol{ \beta}_i - {\mathbf{\Psi}}^\textsf{T}\,{\textbf{g}(t_i)}\|^2} 
  \end{split}
  \label{Eq:DA1}
\end{equation}
\normalsize
 where, 
\small$\mathbf{G}\,\,$\normalsize is an \small$N_{sc} \times N_{sc}\,$\normalsize symmetric matrix, such that both the \small$i^{th}\,$\normalsize   row and the \small$i^{th}\,$\normalsize column are equal to \small$\textbf{g}(t_i)$\normalsize, \small$i=1: N_{sc}$\normalsize; \small$\mathbf{K}\,\,$\normalsize  is an \small$N+1 \times N\,$\normalsize matrix, such that the \small$i^{th}\,$\normalsize column is equal to \small$\textbf{k}(x_i)$\normalsize, \small$x_i, i=1:N$\normalsize.
\small$c_k$\normalsize's are loss functions over the constraints defined as
  \small$c_k(\mathbf{G}\, {\mathbf{\Psi}}\, \mathbf{K})) = (max(0, (l-\textbf{1}_i^\textsf{T} \mathbf{G}\, {\mathbf{\Psi}}\, \mathbf{K} \textbf{1}_j) ))^2\,$\normalsize for same class pairs of index \small$i\,$\normalsize and \small$j$\normalsize,  or \small$ =r\cdot(max(0, (\textbf{1}_i^\textsf{T} \mathbf{G}\, {\mathbf{\Psi}}\, \mathbf{K} \textbf{1}_j -u) ))^ 2\,$\normalsize otherwise, where \small$\textbf{1}_i\,$\normalsize is an \small$N_{sc} \times 1\,$\normalsize vector with all zeros except at index \small$i$\normalsize, \small$\textbf{1}_j\,$\normalsize is an \small$N \times 1\,$\normalsize vector with all zeros except at index \small$j$\normalsize. This leads to that   \small$ c_k(\mathbf{G}\, {\mathbf{\Psi}}\, \mathbf{K})) = (max(0, (l-\textbf{g}(t_i)^\textsf{T}\, {\mathbf{\Psi}}\, \textbf{k}(x_j) ))^2\,$\normalsize for same class pairs of index \small$i\,$\normalsize and \small$j$\normalsize, or \small$ =r\cdot(max(0, (\textbf{g}(t_i)^\textsf{T}\, {\mathbf{\Psi}}\,\textbf{k}(x_j) -u) ))^ 2$\normalsize otherwise, where \small$u>l$\normalsize\ignore{ (note any appropriate $l$, $u$ could work in our case we used $l =2$, $u=-2$ )}, \small$r = \frac{nd}{ns}\,$\normalsize such that \small$nd\,$\normalsize and \small$ns\,$\normalsize are the number of pairs \small$(i,j)\,$\normalsize of different classes and similar pairs respectively. \ignore{ \small$r(\cdot)\,$\normalsize is a matrix regularizer;} Finally, we used a Frobenius norm regularizer for \small$r({\mathbf{\Psi}})$\normalsize.

The objective function in Eq ~\ref{Eq:DA1}  controls the involvement of the constraints \small$c_k\,$\normalsize by the term multiplied by \small$\lambda_1$\normalsize, which controls its importance; we call it \small$C_{l,u}({\mathbf{\Psi}})$\normalsize. While, the trained classifiers penalty is captured by the term multiplied by \small$\lambda_2$\normalsize; we call it \small$C_{\beta}({\mathbf{\Psi}})$\normalsize. One important observation on  \small$C_{\beta}({\mathbf{\Psi}})$\normalsize, is that it reaches zero when \small${\mathbf{\Psi}} = \mathbf{G}^{-1} \textbf{B}^\mathsf{T}$\normalsize, where \small$\textbf{B}  = [\boldsymbol{\beta}_1 \cdots \boldsymbol{\beta}_{N_{sc}}]$\normalsize, since it could be rewritten as \small$C_{\beta}({\mathbf{\Psi}}) = \|\textbf{B}^\mathsf{T} - \mathbf{G} \, {\mathbf{\Psi}}  \|_{F}^2$\normalsize.\ignore{ Our intuition is that for the model to have good generalization, the effect of \small$C_{\beta}({\textbf{T}})\,$\normalsize should be  minimal (\ie \small$\lambda_2 \to 0$\normalsize), since this case indicates successful modeling of the transfer from \small$\mathcal{E}\,$\normalsize domain to the kernel-classifier parameters in \small$\mathcal{X}\,$ domain. }
\ignore{
In contrast to the linear-classifier restricted approach proposed by Elhoseiny et al    \cite{Hoseini13}, our domain transfer model can transfer any type of classifier of an arbitrary kernel from $\mathcal{T}$ to $\mathcal{V}$. Furthermore, the classifier penalty term was not studied in \cite{Hoseini13}, which is captured here by $C_{\beta}({\textbf{T}})$.}

We minimize \small$L({\mathbf{\Psi}})\,$\normalsize  by gradient-based optimization using a \ignore{second order BFGS }quasi-Newton optimizer. Our  gradient derivation of \small$L({\mathbf{\Psi}})\,$\normalsize leads to the following form
\small
\begin{equation}
\frac{\partial L({\mathbf{\Psi}})}{\partial \, {\mathbf{\Psi}}} =  {\mathbf{\Psi}} + \lambda_1 \cdot  \sum_{i,j} {\mathbf{g}(t_i)}   {\mathbf{k}(x_j)}^\mathsf{T} v_{ij} +
r \cdot \lambda_2 \cdot ( \mathbf{G}^2\,\, {\mathbf{\Psi}} - \mathbf{G} \textbf{B}) 
\label{eq:grd}
\end{equation}
\normalsize
where \small$v_{ij} = - 2 \cdot max(0, (l-\textbf{g}(t_i)^\mathsf{T}\, {\mathbf{\Psi}}\, \textbf{k}(x_j) )\,$\normalsize if \small$i\,$\normalsize and \small$j\,$\normalsize correspond to the same class, \small$2 \cdot max(0, (\textbf{g}(t_i)^\mathsf{T}\, {\mathbf{\Psi}}\, \textbf{k}(x_j) -u )\,$\normalsize otherwise. Another approach that can be used to minimize \small$L({\mathbf{\Psi}})\,$\normalsize is through alternating projection using Bregman algorithm \cite{bregman97}, where \small${\mathbf{\Psi}}\,$\normalsize is updated by a single constraint every iteration.

\subsection{Kernel Classifier Prediction}
\label{ss_kmethods}
We study two ways to infer the final kernel-classifier prediction. (1) Direct Kernel Domain Transfer Prediction, denoted by ``DT-kernel'', (2) One-class SVM adjusted DT Prediction, denoted by ``SVM-DT kernel''. Hyper-parameter selection is attached in the supplementary materials. The source code is available here 
\footnotesize{\url{https://sites.google.com/site/mhelhoseiny/computer-vision-projects/write_kernel_classifier}}.\normalsize  

\medskip
\noindent \textbf{Direct Domain Transfer (DT) Prediction:} By construction a classifier of an unseen class can be directly computed from our trained domain transfer model as follows
\small
\begin{equation}
\centering
\begin{split}
\Phi(t_*) = \tilde{\boldsymbol{\beta}}_{DT}(t_*) = {\mathbf{\Psi}^*}^\textsf{T} \, \mathbf{g}(t_*)
\end{split}
\end{equation} 
\normalsize
\medskip
\noindent \textbf{One-class-SVM adjusted DT (SVM-DT) Prediction:} 
In order to increase separability against seen classes, we adopted the inverse of the idea of the one class kernel-svm, whose main idea is to build a confidence function that takes only positive examples of the  class. Our setting is the opposite scenario; seen examples are negative examples of the unseen class.
In order to introduce our proposed adjustment method, we  start by presenting the one-class SVM objective function. The  Lagrangian dual  of the one-class SVM~\cite{oneclasssvm07} can be written as
\small
\begin{equation}
\label{eq:1class}
\small
\begin{split}
{\boldsymbol{\beta}}^*_{+} =  &   \underset{\boldsymbol{\beta}}{\operatorname{argmin }}\big[    \boldsymbol{\beta}^\textsf{T} \mathbf{K^{' }}\boldsymbol{\beta} - \boldsymbol{\beta}^T \mathbf{a} \big]\\
&s.t.: \boldsymbol{\beta}^T \mathbf{1} = 1,  0 \le \boldsymbol{\beta}_i \le C; i = 1 \cdots N   \\
\end{split}
\end{equation}
\normalsize
where \small$\mathbf{K^{' }}\,$\normalsize is an \small$N \times N\,$\normalsize matrix, \small$\mathbf{K^{' }}(i,j) = k({x}_i, {x}_j)$\normalsize, \small$\forall {x}_i,{x}_j \in \mathcal{S}_x$\normalsize (\ie in the training data), \small$\textbf{a}\,$\normalsize is an \small$N \times 1\,$\normalsize vector, \small$\textbf{a}_i = k({x}_i, {x}_i)$\normalsize, \small$C\,$\normalsize is a hyper-parameter . It is straightforward to see that, if $\beta$ is aimed to be a negative decision function instead, the objective function would become in the following form
\small
\begin{equation}
\label{eq:1classneg}
\small
\begin{split}
{\boldsymbol{\beta}}^*_{-} =  &   \underset{\boldsymbol{\beta}}{\operatorname{argmin }}\big[    \boldsymbol{\beta}^\textsf{T} \mathbf{K^{' }}\boldsymbol{\beta} + \boldsymbol{\beta}^T \mathbf{a} \big]\\
&s.t.: \boldsymbol{\beta}^T \mathbf{1} = -1, -C \le \boldsymbol{\beta}_i \le 0; i = 1 \cdots N \\
\end{split}
\end{equation}
\normalsize
While \small${\boldsymbol{\beta}}^*_{-}  = - {\boldsymbol{\beta}}^*_{+}$\normalsize, the objective function in Eq~\ref{eq:1classneg} of the one-negative class SVM inspires us with the idea to adjust the kernel-classifier parameters to increase separability of the unseen kernel-classifier against the points of the seen classes, which leads to the following objective function 
\small
\begin{equation}
\label{eq:form_kernel}
\small
\begin{split}
\Phi(t_*) = \hat{\boldsymbol{\beta}}(t_*) =  &   \underset{\boldsymbol{\beta}}{\operatorname{argmin }}\big[    \boldsymbol{\beta}^\textsf{T} \mathbf{K^{' }}\boldsymbol{\beta} - \zeta \hat{\boldsymbol{\beta}}_{DT}(t_*)^\textsf{T} \mathbf{K^{'}} \boldsymbol{\beta}   + \boldsymbol{\beta}^T \mathbf{a} \big]\\
&s.t.:   \boldsymbol{\beta}^T \mathbf{1} = -1, {\hat{\boldsymbol{\beta}}_{DT}}^{\mathsf{T}} \mathbf{K^{' }} \boldsymbol{\beta}> l, -C \le \boldsymbol{\beta}_i \le 0;\forall i \\
& C, \zeta , l \,  \text{: hyper-parameters},\\ 
\end{split}
\end{equation}
\normalsize
where  \small$\hat{\boldsymbol{\beta}}_{DT}\,$\normalsize is the first \small$N\,$\normalsize elements in \small$\tilde{\boldsymbol{\beta}}_{DT}(t^*) \in \mathbb{R}^{N+1}$\normalsize, \small$\mathbf{1}\,$\normalsize is an \small$N \times 1\,$\normalsize vector of ones. The objective function, in Eq~\ref{eq:form},  pushes the classifier of the unseen class to be highly correlated with the domain transfer prediction of the kernel classifier, while putting  the points of the seen classes as negative examples. It is not hard to see that Eq~\ref{eq:form_kernel} is a quadratic program in \small$\mathbf{\beta}$\normalsize, which could be solved using any quadratic solver.
\ignore{In contrast to our formulation, the approaches presented in~\cite{NIPS13DeViSE,NIPS13CMT,Hoseini13} assumes that \small$\mathcal{X} \in R^{d_b}$ and  $\mathcal{E} \in R^{d_E}\,$\normalsize (\ie  vectorized).} It is worth to mention that linear classifier prediction in Eq~\ref{eq:form} (best Linear formulation in our results)  predicts  classifiers by solving an optimization problem of size  \small$N+d_v+1\,$\normalsize  variables, \small$d_v+1\,$\normalsize linear-classifier parameters\ignore{, which is the same as the length of the visual feature vector,} and \small$N\,$\normalsize slack variables\ignore{; a similar limitation can be found in~\cite{NIPS13DeViSE,NIPS13CMT} where the architecture depends on the number on visual features}.  In contrast, the kernelized objective function (Eq~\ref{eq:form_kernel}) solves a  quadratic program of only \small$N\,$\normalsize variables, and  predicts a kernel-classifier instead with fewer parameters. Using very high-dimensional features  will not affect the optimization complexity.  \ignore{
Therefore, it is clear that the kernel formulation is expected to have better generalization properties. In addition, the kernel-approach does not assume that any of \small$\mathcal{V}\,$\normalsize and \small$ \mathcal{T}\,$\normalsize is a vector space.}

\section{Distributional Semantic (DS) Kernel for text  descriptions}
\label{dskernel}

We propose a distributional semantic kernel $g(\cdot, \cdot) = g_{DS}(\cdot, \cdot)$  to define the similarity between two text descriptions in $\mathcal{T}$ domain\ignore{of visual classes in our setting}. While this kernel is applicable to  kernel classifier predictors  presented in Sec~\ref{sec:app}, it could be used for other applications. We start by  distributional semantic models in~\cite{mikolov2013distributed,mikolov2013efficient} to represent the semantic manifold $\mathcal{M}_s$, and a function $vec(\cdot)$ that maps a word to a $K\times 1$ vector in $\mathcal{M}_s$. The main assumption behind this class of distributional semantic model  is that similar words share similar context. Mathematically speaking, these models  learn a vector for each word $w_n$, such  that $p(w_n|(w_{n-L}, w_{n-L+1}, \cdots,  w_{n+L-1},w_{n+L})$ is maximized over the training corpus, where $2\times L$ is the context window size. Hence similarity between $vec(w_i)$ and $vec(w_j)$ is high if they co-occurred a lot in context of size $2\times L$ in the training text-corpus. We normalize all the word vectors to length $1$ under L2 norm, i.e., $\| vec(\cdot) \|^2=1$. 

Let us assume a  text description ${D}$ that we represent by a set of triplets ${D} = \{(w_l,f_l, vec(w_l)), l=1\cdots M\}$, where $w_l$ is a word that occurs in ${D}$ with frequency $f_l$ and its corresponding word vector is $vec(w_l)$ in $\mathcal{M}_s$. We drop the stop words from ${D}$. We define  $\textbf{F} = [f_1, \cdots, f_M]^\textsf{T}$ and $\textbf{P} = [vec(w_1), \cdots, vec(w_M)]^\textsf{T}$, where $\textbf{F}$ is an $M\times1$  vector of term frequencies and $\textbf{P}$ is an $M \times K$ matrix of the corresponding term vectors. 

Given two text descriptions ${D}_i$ and ${D}_j$ which contain $M_i$ and $M_j$ terms respectively. We compute $\textbf{P}_i$ ($M_i \times 1$) and $\textbf{V}_i$ ($M_i \times K$) for  ${D}_i$  and $\textbf{P}_j$ ($M_j \times 1$) and $\textbf{V}_j$ ($M_j \times K$) for  ${D}_j$. Finally  $g_{DS}({D}_i, {D}_j)$ is defined as 
\begin{equation}
g_{DS}({D}_i, {D}_j) = \textbf{F}_i^\textsf{T} \textbf{P}_i \textbf{P}_j^\textsf{T}  \textbf{F}_j
\end{equation}
One advantage of this similarity measure is that it captures semantically related terms. It is not hard to see that the standard Term Frequency (TF) similarity could be thought of as a special case of this kernel where $vec(w_l)^\mathsf{T} vec(w_m)=1$ if $w_l=w_m$, 0 otherwise, i.e., different terms are orthogonal. However, in our case the word vectors are learnt through a distributional semantic model which makes semantically related terms have higher dot product ($vec(w_l)^\mathsf{T} vec(w_m)$).

\section{Experiments}
\label{experiments}


\subsection{Datasets and Features}
\label{ss_ds_feats}
\noindent{\bf Datasets:}
We evaluated our methods using two large datasets, widely used for fine-grained categorization:  CU200 Birds~\cite{CU20010}  dataset (200 classes - 6033 images) and the Oxford Flower-102~\cite{Flower08} dataset (102 classes - 8189 images). 
We augmented these datasets with a textual description of each category. 
The CUB200 Birds image dataset was created based on birds that have a corresponding Wikipedia article, so we have developed a tool to automatically extract Wikipedia articles given the class name. The tool succeeded to automatically generate 178 articles, and the remaining 22 articles was extracted manually from Wikipedia. These mismatches happen when article title is a different synonym of the same bird class. On the other hand, for Flower dataset, the tool managed to generate only 16 classes from Wikipedia out of 102 since the Flower  classes do not necessarily have corresponding Wikipedia articles. The remaining 86 articles were generated manually for each class from Wikipedia, Plant Database  \footnote{http://plants.usda.gov/java/}, Plant Encyclopedia \footnote{http://www.theplantencyclopedia.org/wiki/Main$\_$Page}, and BBC articles \footnote{http://www.bbc.co.uk/science/0/}. The collected textual descriptions for Flowers and Birds datasets are available here  \url{https://sites.google.com/site/mhelhoseiny/1202-Elhoseiny-sup.zip} .

\noindent{\bf Textual Feature Extraction:}
The textual features were extracted in two phases.
     The first phase is an indexing phase that generates textual features with tf-idf (Term Frequency-Inverse Document Frequency) configuration (Term frequency as local weighting while inverse document frequency as a global weighting). The tf-idf is a measure of how important  a word is to a text corpus. The tf-idf value increases proportionally to the number of times a word appears in the document, but is offset by the frequency of the word in the corpus, which helps to control for the fact that some words are generally more common than others. We used the normalized frequency of a term in the given textual description~\cite{salton1988term}. The inverse document frequency is a measure of whether the term is common; in this work we used the standard logarithmic idf~\cite{salton1988term}.   
The second phase is a dimensionality reduction step, in which  Clustered Latent Semantic Indexing (CLSI) algorithm~\cite{clsi05} is used. CLSI is a low-rank approximation approach for dimensionality reduction, used for document retrieval. In the Flower Dataset, tf-idf features $\in \mathbb{R}^{8875}$ and after CLSI the final textual features $\in \mathbb{R}^{102}$. In the Birds Dataset, tf-idf features is in $\mathbb{R}^{7086}$ and after CLSI the final textual features is in $\mathbb{R}^{200}$. 

\noindent{\bf  Visual features Extraction:}
We used the Classemes features~\cite{classemes} as the visual feature for our experiments, where they provide an intermediate semantic representation of the input image. Classemes features are output of a set of classifiers corresponding to a set of $C$ category labels, which are drawn from an appropriate term list defined in~\cite{classemes}, and not related to our textual features. For each category $c \in \lbrace 1 \cdots C \rbrace $, a set of training images is gathered by issuing a query on the category label to an image search engine.
After a set of coarse feature descriptors (Pyramid HOG, GIST, \etc) is extracted, a subset of feature dimensions was selected~\cite{classemes}, and  a one-versus-all classifier $\varphi_{c}$ is trained for each category. The classifier output is real-valued, and is such that $\varphi_{c}(x) > \varphi_{c}(y)$ implies that $x$ is more similar to class $c$ than $y$ is. Given an image $x$,  the feature vector (descriptor) used to represent it is the Classemes vector $  [\varphi_{1} (x),  \cdots, \varphi_{d_v} (x)]$, $d_v=2569$.

For Kernel classifier prediction, we evaluated  these features and also additional representations  for text descriptions and images. For text, we performed experiments with the proposed distributional semantic kernel and using Recurrent Nets. For images, we evaluated (a) CNN features and (b), combined kernel over different features learnt by MKL (multiple kernel learning)). Details are discussed later in Subsection~\ref{ss_exp_kernel}.


\begin{figure*}[ht!]
\centering
\hspace{-10mm}
\begin{minipage}{0.33\textwidth}
  \centering
\includegraphics[width=1.1\linewidth,height=.75\linewidth]
{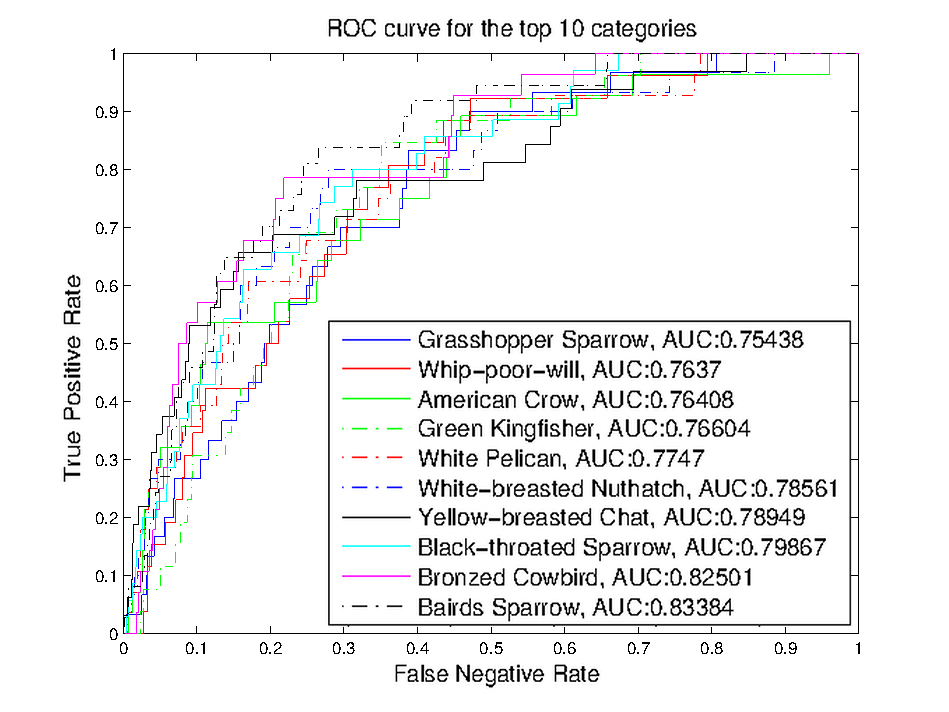}
\end{minipage}%
\hspace{-3mm}
\begin{minipage}{0.33\textwidth}
  \centering
 \includegraphics[width=1.1\linewidth,height=.75\linewidth]
{birds_top10_ROC_final2.jpg}
\end{minipage}
\begin{minipage}{0.33\textwidth}
  \centering
\includegraphics[width=1.1\linewidth,height=.75\linewidth]{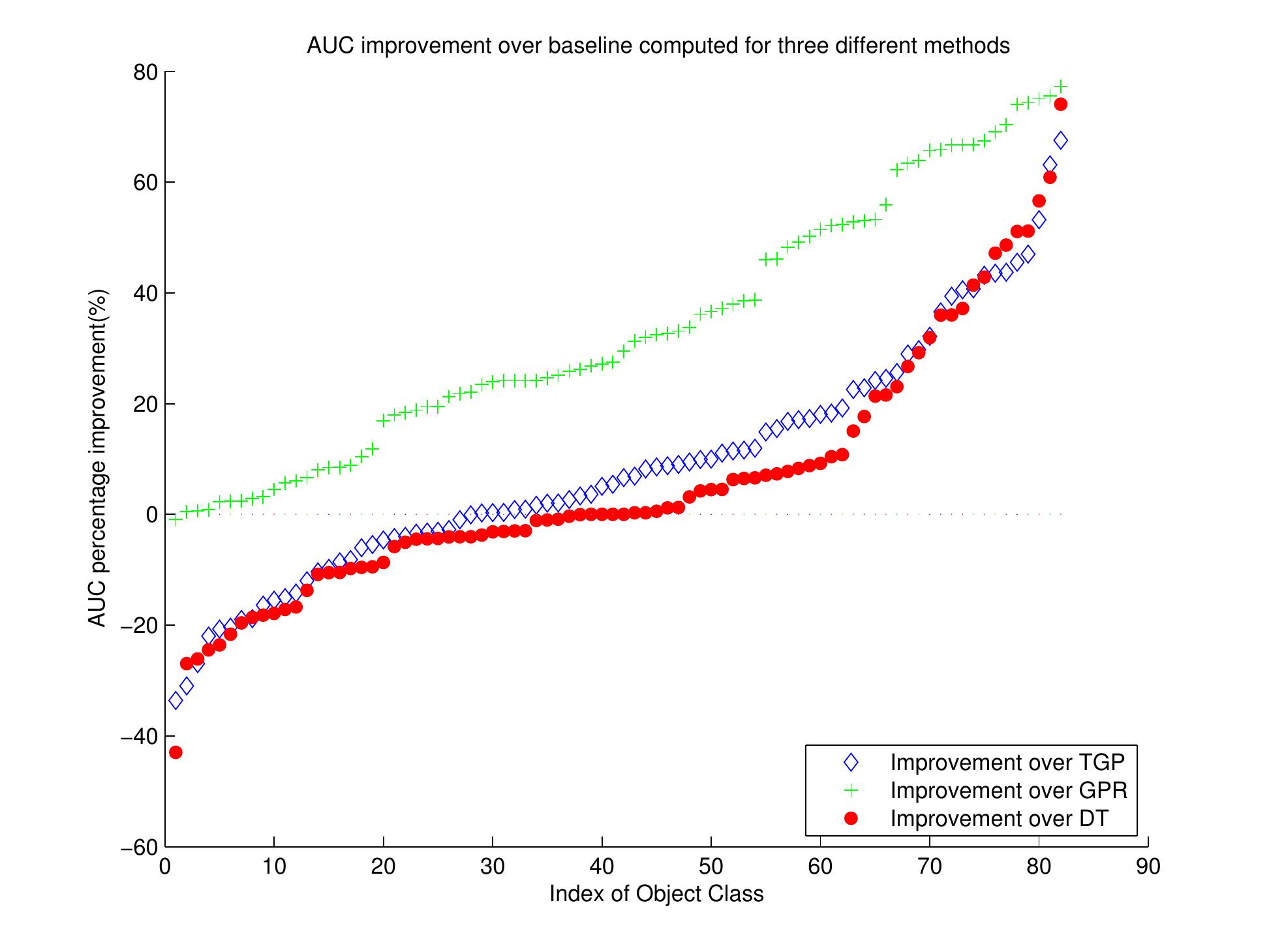}
\end{minipage}
\vspace{-2mm}
\caption{Linear : \textbf{Left and Middle}: ROC curves of best 10 predicted classes by the final formulation (E) for Bird  and Flower datasets respectively, \textbf{Right}: AUC improvement over the three baselines on Flower dataset (Formulations A (GPR), A (TGP), C). The improvement is sorted in an increasing order for each baseline separately (best seen in color)}
\vspace{-2mm}
\label{fig:result3fig}
\end{figure*}
\begin{table*}[b!]
\small
\centering
\vspace{-2mm}
\caption{ Linear: Comparative Evaluation of Different Formulations on the Flower and Bird Datasets}
\label{T:AUC}
\scalebox{1.0}{
\begin{tabular}{|l|l|l|}
  \hline
  		&  Oxford Flowers   & UC-UCSD Birds   \\
  Approach  & Avg AUC (+/- std) & Avg AUC (+/- std)\\ 
  \hline
  \hline
  (A) Regression - GPR & 0.54 (+/- 0.02) & 0.52 (+/- 0.001) \\ 
  (A) Structured Regression - TGP & 0.58 (+/- 0.02) & 0.61 (+/- 0.02)\\
  (C)  Domain Transfer(DT) &  0.62(+/- 0.03)  &  0.59 (+/- 0.01)\\
  \hline 
  (B) Constrained GPR & 0.62(+/- 0.005) & - \\
  (B) Constrained TGP & 0.63(+/- 0.007) & - \\
  (D) Constrained Domain Adaptation (CDT) on Eq~\ref{eq:form} & 0.64 (+/- 0.006)& -  \\
       \hline
  (E) Regression+DT + constraints (final best linear approach) & {0.68} (+/- 0.01) & { 0.62} (+/- 0.02) \\
      \hline
\end{tabular}
\vspace{-2mm}
}
\ignore{
{
\scalebox{0.7}{
\begin{tabular}{|l|l|l|l|l|}
  \hline
  \multicolumn{5}{c}{Top-5 Classes with highest combined improvement} \\
  \hline
  \hline
  class      &  (A) TGP (AUC) & (C) DT (AUC) & (D) TGP+DT+C  & \% Improv. \\
  \hline
  \hline
   2   &  0.51 & 0.55 & 0.83 & 57\% \\  
   28 & 0.52 & 0.54 & 0.76 &  43.5\% \\
   26 &  0.54 & 0.53 & 0.76 & 41.7\% \\
   81 & 0.52 & 0.82 & 0.87   & 37\%  \\
   37 & 0.72 & 0.53 & 0.83   & 35.7 \% \\
  \hline 
\end{tabular}}
}}
\end{table*}

\subsection{Experimental Results for Linear Classifier Prediction}

\noindent{\bf Evaluation Methodology:} Following  zero-shot learning literature, we evaluated the performance of an unseen classifier in a one-vs-all setting where the test images of unseen classes are considered to be the positives and the test images from the seen classes are considered to be the negatives. We computed the ROC curve and report the area under that curve (AUC) as a comparative measure of different approaches. In zero-shot learning setting the test data from the seen classes are typically very large compared to those from unseen classes. This makes other measures, such as accuracy, useless since high accuracy can be obtained even if all the unseen class test data are  classified incorrectly; hence we used ROC curves, which are independent of this problem.  

\noindent{\bf  Training/Testing ZSL Splits}

\noindent \textit{\textbf{Super Category Unseen (SC-Unseen) Split)}}. This is  Zero-shot  setting split for both CUB and Flower Datasets (first defined in our work~\cite{Hoseini13}).  Five-fold cross validation over the classes were performed, where in each fold 4/5 of the classes are considered as ``seen classes'' and are used for training and 1/5th of the classes were considered as ``unseen classes'' where their classifiers are predicted and tested. Within each of these class-folds, the data of the seen classes are further split into training and test sets. The hyper-parameters for the  approach were selected through another five-fold cross validation within the class-folds  (i.e. the $80\%$ training classes are further split into $5$ folds to select the hyper-parameters). We made the seen-unseen folds used in our experiments available here \url{https://sites.google.com/site/mhelhoseiny/computer-vision-projects/Write_a_Classifier}. In contrast to  the SC-seen split, discussed next,  this  split was designed such that bird subspecies that belong to the same super-category should either belong to either the training or the test split. 

\noindent  \textit{\textbf{Super Category Seen (SC-Seen) Split} a (150-50) Split on CUB 2011 dataset~\cite{akata2015evaluation}:}  We also evaluate our work on another zero-shot learning split for CUB 2011 dataset, which is used in some recent works  (e.g,~\cite{akata2015evaluation,qiao2016less}). \ignore{This split was not available at the time this manuscript was written. }We investigated the difference between this training/testing split and found that most of the unseen/test classes in split defined in ~\cite{akata2015evaluation} are actually seen in some-perspective. In particular, we found a common feature in this split is that for each group of related subordinate categories, the majority of the group subspecies is used during training and one of them is left as unseen. For instance,  all subspecies of Albatrosses  are included among the  training classes except one  for testing (i.e., training on  \texttt{Laysan\_Albatross} and \texttt{Sooty\_Albatross},  and testing on  \texttt{Black\_footed\_Albatross}). At test time, a zero-shot learning model is asked to  discriminate  between \texttt{Black\_footed\_Albatross } and other classes that are not related to Albatross which is relatively easier given that the model has seen already Albatrosses during training. Hence, we name this split Super Category Seen(SC-Seen) Split. Instead in our Super Category Unseen (SC-Unseen) Split, the whole set of albatrosses and other unseen subordinate categories  are completely unseen and at test time the model is asked to discriminate between different types of Albatrosses from just their text. This make the SC-Unseen split much more difficult than SC-Seen split. All of our CUB dataset was based on 2010 version (with 6033 images) and on the SC-Unseen split and Wikipedia Articles from 2012. In order to show our results in comparison with some recently published work, we applied our methods on the SC-Seen Split discuss our findings in Sec~\ref{ss_soa}).

\noindent{\bf Baselines:}
Since our work is the first to predict classifiers based on pure textual description, there are no other reported results to compare against. However, for further comparisons we designed three state-of-the-art baselines to compare against, which are designed to be inline with our argument in Sec~\ref{obvformulation}. Namely we used: 1) A Gaussian Process Regressor (GPR)~\cite{Rasmussen:2005}, 2) Twin Gaussian Process (TGP)~\cite{Bo:2010} as a structured regression method, 3)  Domain Transfer (DT)~\cite{da11}. The TGP and DT baselines are of particular importance since they are incorporated in our formulation. 
It has to be noted that we also evaluate TGP and DT as alternative formulations that we are proposing for the problem, none of them was used in the same context before. 

\begin{figure}
\vspace{-3mm}
\centering
\includegraphics[width=1.01\linewidth,height=.27\linewidth]{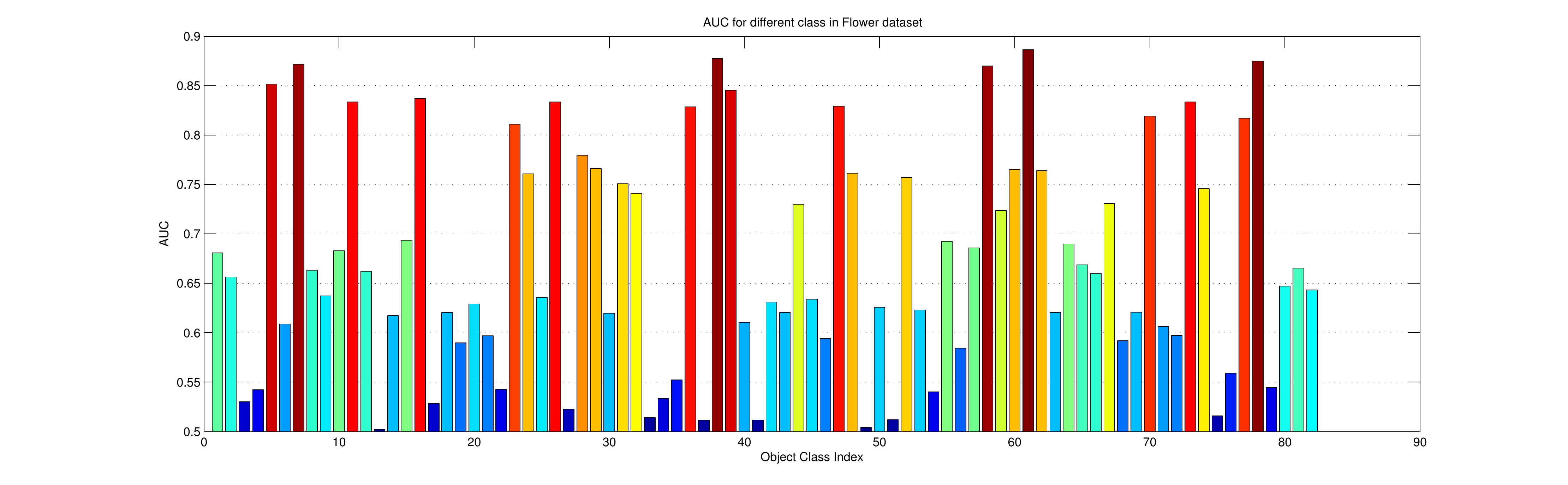}
\vspace{-3mm}
\caption{Linear: AUC of the predicated classifiers for all classes of the flower datasets (Formulation E)}
\label{F:AUCs}
\vspace{-2mm}
\end{figure}

\noindent{\bf Results:}
Table~\ref{T:AUC} shows the average AUCs for the final linear approach in comparison to the three baselines on both datasets. GPR performed poorly in all classes in both data sets, which was expected since it is not a structure prediction approach.  The DT formulation outperformed TGP in the flower dataset but slightly underperformed on the Bird dataset. The proposed approach outperformed all the baselines on both datasets, with significant improvement on the flower dataset. It is also clear that the TGP performance was improved on the Bird dataset since it has more classes (more points are used for prediction). Fig~\ref{fig:result3fig} shows the ROC curves for our approach on best predicted unseen classes from the Birds dataset on the Left  and Flower dataset on the middle. Fig ~\ref{F:AUCs} shows the AUC for all the classes on Flower dataset. 

\begin{table}[t]
\small
\caption{\small Linear: Percentage of classes that the final proposed approach (formulation (E)) makes an improvement in predicting over the baselines (relative to the total number of classes in each dataset)}
\label{T:classimprov}
\centering
\scalebox{1.0}{
\begin{tabular}{|l|l|l|}
  \hline
  		&  Flowers (102)  & Birds (200)\\ 
 baseline      &  \%  improvement & \%  improvement\\  
  \hline
  \hline
  (A) GPR  & 100 \% & 98.31 \% \\ 
  (A) TGP  & 66 \% & 51.81 \%\\
  (C) DT  &   54\% &  56.5\% \\
  \hline 
\end{tabular}}
\end{table}
Fig ~\ref{fig:result3fig}, on the right, shows the improvement over (A) GPR, A(TGP), and (C) DT for each class, where the improvement is calculated as (our AUC- baseline AUC)/ baseline AUC \%. 
Table~\ref{T:classimprov} shows the percentage of the classes which our approach makes a prediction improvement for each of  the three baselines
. Table~\ref{T:top5improv} shows the five classes in Flower dataset where our approach made the best average improvement.
This table shows that in these cases both TGP and DT performed poorly while our formulation that is based on both of them has a significant improvement. This shows that our formulation does not simply combine the best of the two approaches but can significantly improve the prediction performance.

\begin{table}[t]
\small
\caption{\small Linear: Top-5 classes with highest combined improvement in Flower dataset}
\label{T:top5improv}
\centering
{
\scalebox{0.82}{
\begin{tabular}{|l|l|l|l|l|}
  \hline
  class      &  (A) TGP (AUC) & (C) DT (AUC) & (E) Our (AUC)  & \% Improv. \\
  \hline
  \hline
   2   &  0.51 & 0.55 & 0.83 & 57\% \\  
   28 & 0.52 & 0.54 & 0.76 &  43.5\% \\
   26 &  0.54 & 0.53 & 0.76 & 41.7\% \\
   81 & 0.52 & 0.82 & 0.87   & 37\%  \\
   37 & 0.72 & 0.53 & 0.83   & 35.7 \% \\
  \hline 
\end{tabular}}
}
\vspace{-3mm}
\end{table}

To evaluate the effect of the constraints in the objective function in Eq~\ref{eq:form}, we removed the constraints $- (\mathbf{c}^\textsf{T} {\mathbf{x}}_{i} ) \geq \zeta_i$ which enforces all the seen examples to be on the negative side of the predicted classifier hyperplane and evaluated the approach. The result on the flower dataset (using one fold) was reduced to average AUC=0.59 compared to AUC=0.65 with the constraints. Similarly, we evaluated the effect of the constraint  $\mathbf{t}_*^\textsf{T} \mathbf{W} \mathbf{c} \ge l$. The result was reduced to average AUC=0.58 compared to AUC=0.65 with the constraint.  This illustrates the importance of this constraint in the formulation.

\noindent{\bf Constrained Baselines:}
\ignore{We computed the ROC curves and report the area under that curve (AUC) as a comparative measure\footnote{In zero-shot learning setting the test data from the seen class are typically very large compared to those from unseen classes. This makes other measures, such as accuracy, useless since high accuracy can be obtained even if all the unseen class test data are wrongly classified; hence we used ROC curves, which are independent of this problem.} Five-fold cross validation over the classes were performed, within each of these class-folds, the data of the seen classes are further split into training and test sets.}
Table~\ref{T:AUC} also shows the average AUCs for the constrained baseline formulations, namely (B) Constrained GPR Regression, (B) Constrained TGP Regression and (D) Constrained DT; see section~\ref{formulation}. 
As  previously discussed, GPR performed poorly, while, as expected, TGP performed better. Adding constraints to GPR/TGP improved their performance. Combining regression and DT gave significantly better results for classes where both approaches individually perform poorly, as can be seen in Table~\ref{T:AUC}. We performed an additional experiment, where  $\mathbf{W}$ is computed using Constrained Domain Transfer (CDT). Then, the unseen classifier is predicted using equation~\ref{eq:form} with $\gamma=0$, which performs worse. This indicates that adding constraints to align to seen classifiers hurts the learnt domain transfer function on unseen classes. In conclusion, the final formulation (Eq~\ref{eq:form}) that combines TGP and DT with additional constraints performs the best in both Birds and Flower datasets. The effect of TGP is very limited since it was trained on sparse points which is reflected in the setting of $\alpha$ (weight for DT) and $\gamma$ (weight for TGP) to $100$ and $1$ respectively after hyper parameter tuning on a validation set. 


\subsection{Experimental Results for Kernel Classifier Prediction}
\label{ss_exp_kernel}

\ignore{
Now that we have described our zero-shot learning setting and the suggested approaches to directly predict kernel-classifier parameters for unseen classes, we present several experiments to validate our model.
}
\ignore{In this section, we presented a set of experiments, conducted to evaluate our proposed model for zero-shot learning of visual classifiers. The quantitative comparisons show our superior performance to the state of the art on two challenging datasets of fine-grained object categories.}


\subsubsection{Additional Evaluation Metrics}

In addition to the AUC, discussed in the previous section, we report two additional metrics while evaluating and comparing the kernel classifier prediction to the linear classifier prediction, detailed as follows


\textit{\small$|N_{sc}|\,$\normalsize to \small$|N_{sc}+1|\,$\normalsize Recall:}
 this metric check  how   the learned classifiers of the seen classes confuse the predicted classifiers, when they are involved in a multi-class classification problem of \small$N_{sc} + 1\,$\normalsize classes. \ignore{The first \small$N_{sc}\,$\normalsize classifiers are those of the seen classes, while \small$({N_{sc}+1})^{st}$\normalsize classifier is a predicted classifier for an unseen class. }We use Eq~\ref{eq:mclass} to predict label $l^*$  with the maximum confidence of an image \small$x^*$\normalsize, such that \small$l^* \in {L}_{sc} \cup l_{us}$\normalsize,  \small$l_{us}\,$\normalsize is the label of the ground truth unseen class, and ${L}_{sc}$ is the set of seen class labels. We compute the recall under this setting. This metric is computed for each predicted unseen classifier and the average is reported.

\textit{Multiclass Accuracy of Unseen classes (MAU):} Under this setting, we aim to evaluate the performance of  the unseen  classifiers against each others. Firstly, the classifiers of all unseen categories are predicted. Then, we use Eq~\ref{eq:mclass} to predict the label with the maximum confidence of a test image $x$, such that its label $l_{us}^* \in {L}_{us}$, where ${L}_{us}$ is the set of all unseen class labels that only have text descriptions.



\subsubsection{Comparisons to Linear Classifier Prediction}
\label{exp1}
 We compare the kernel methods to the linear prediction discussed earlier,  which predicts a linear classifier from textual descriptions  ( \small$\mathcal{T}\,$\normalsize space in our framework). The goal is to check whether the predicted kernelized classifier outperforms the predicted linear classifier. 
We used the same features on the visual domain  and the textual domains detailed in subsection~\ref{ss_ds_feats}. 
 
We denote our kernel Domain Transfer prediction  and one class SVM adjusted DT prediction presented  in Section~\ref{ss_kmethods} by  ``DT-kernel'' and ``SVM-DT-kernel'' respectively. We compared against  linear classifier prediction (Linear Formulation (E) approach, denoted by just Linear Classifier).  \ignore{(which uses a quadratic program to optimize the classifier parameters)}We also compared against the  linear direct domain transfer (Linear Formulation (C), denoted by DT-linear).  In our kernel approaches, we used Gaussian rbf-kernel as a similarity measure in \small$\mathcal{T}\,$\normalsize and \small$\mathcal{V}\,$\normalsize spaces (\ie \small$k(d,d') = exp(-\lambda ||d-d'||)$\normalsize).

\begin{table*}[hb!]
\vspace{-1mm}
 \begin{minipage}{0.38\linewidth}
\caption{Kernel: Recall, MAU, and average AUC on three seen/unseen splits on Flower Dataset and a seen/unseen split on Birds dataset}
\label{tbl:flowerbirdsmauauc}
  \centering
  \scalebox{0.90}
  {
\begin{tabular}{|c|c|c|}
\hline 
  & \textbf{Recall-Flower}  & \textbf{Recall-Birds} \\ 
  \hline 
{SVM-DT kernel-rbf }& \textbf{40.34\% (+/-  1.2) \%} &  \textbf{44.05 \%}    \\ 
\hline 
Linear Classifier  & 31.33  (+/-  2.22) \%  & 36.56 \%  \\ 
\hline 
\end{tabular}}

\hspace{-1.5mm}\scalebox{0.943}
  {
\begin{tabular}{|c|c|c|}
\hline 
  & \textbf{MAU-Flower}  & \textbf{MAU-Birds} \\ 
  \hline 
{SVM-DT kernel-rbf }& \textbf{9.1 (+/-  2.77) \%}  & \textbf{3.4  \%}    \\ 
\hline 
{DT kernel-rbf }& \textbf{6.64 (+/-  4.1) \%}   & \textbf{2.95  \%} \\ 
\hline 
Linear Classifier \ignore{ Prediction} & 5.93  (+/-  1.48)\%  & 2.62 \% \\ 
\hline 
DT-linear\ignore{~\cite{Hoseini13,da11}} & 5.79 (+/-  2.59)\% &  2.47 \%  \\ 
\hline
\textit{Acc (all classes seen) }&  50.7\% & 16.0\% \\
\hline 
\end{tabular}}
\scalebox{0.93}
{
\begin{tabular}{|c|c|c|} 
\hline 
  & \textbf{AUC-Flower} & \textbf{AUC-Birds}  \\ 
\hline 
{SVM-DT kernel-rbf } & {0.653 (+/-  0.009) }   &   0.61   \\ 
\hline 
{DT kernel-rbf } & {0.623 (+/-  0.01) \%}  &  0.57  \\ 
\hline 
Linear Classifier \ignore{Prediction} & 0.658 (+/-  0.034)  & 0.62  \\ 
\hline 
Domain Transfer\ignore{~\cite{Hoseini13,da11}} & 0.644 (+/-  0.008) &  0.56   \\ 
\hline 
\end{tabular} 
}
\end{minipage}
 \begin{minipage}{00.01\linewidth}
 $\,\,$
 \end{minipage}
\begin{minipage}{0.30\linewidth}
\centering
   \caption{Kernel: MAU on a seen-unseen split-Birds Dataset (MKL)}
\label{tbl:birdsmkl}
 \scalebox{0.93}
  {
\begin{tabular}{|c|c|}
\hline 
& MAU  \\ 
\hline 
{SVM-DT kernel-rbf (text)}& \textbf{4.10 \%}    \\ 
\hline 
Linear Classifier \ignore{Prediction} & 2.74  \\ 
\hline 
\end{tabular} }

\centering
     \vspace{5mm}
   \caption{Kernel: MAU on a seen-unseen split-Birds Dataset (CNN image features, text description)}
\label{tbl:birdscnn}
 \scalebox{0.88}
  {
\begin{tabular}{|c|c|}
\hline 
& MAU  \\ 
\hline 
\hline 
{SVM-DT kernel ($\mathcal{V}$-rbf, $\mathcal{T}$-DS kernel)}& \textbf{5.35  \%}   \\ 
\hline
{SVM-DT kernel ($\mathcal{V}$-rbf, $\mathcal{T}$-rbf on TFIDF)}& \textbf{4.20  \%} \\ 
\hline 
Order Embedding~\cite{vendrov2016order}  & 3.3 \% \\ 
\hline
Linear Classifier (TFIDF text) \ignore{Prediction} & 2.65 \% \\ 
\hline 
\cite{norouzi2014zero} & 2.3\% \\
\hline
\textit{Acc (all classes seen)} & 45.6\% \\
\hline 
\end{tabular} }
\end{minipage}
 \begin{minipage}{00.02\linewidth}
 $\,\,$
 \end{minipage}
\begin{minipage}{0.27\linewidth}
\centering
 \caption{Kernel: Recall and MAU on a seen-unseen split-Birds Dataset (Attributes)} 
\label{tbl:birds3}
 \vspace{-1mm}
\scalebox{1.0}
  {
\begin{tabular}{|c|c|}
\hline 
 & Recall  \\ 
\hline 
{SVM-DT kernel-rbf }& \textbf{76.7  \%}    \\ 
\hline 
Lampert DAP  {\cite{Lampert09}} & 68.1  \%  \\ 
\hline 
\end{tabular}
 }

    \scalebox{1.0}
  {
\begin{tabular}{|c|c|}
\hline 
 & MAU  \\ 
\hline 
{SVM-DT kernel-rbf }& \textbf{5.6  \%}   \\ 
\hline 
{DT kernel-rbf }& {4.03 \%} \\ 
\hline 
Lampert DAP {  \cite{Lampert09}} & 4.8  \%  \\ 
\hline 
\end{tabular}}
\end{minipage}
\vspace{-4mm}
\end{table*}

\textit{Recall metric : } The recall of the SVM-DT kernel approach is 44.05\% for Birds dataset and 40.34\% for Flower dataset, while it is 36.56\% for Birds and  31.33\% for Flower by best Linear Classifier prediction (E). This indicates that the  predicted classifier is less confused by the classifiers of the seen categories compared with  Linear Classifier prediction; see table ~\ref{tbl:flowerbirdsmauauc} (top part)

\textit{MAU metric:} It is worth to mention that the multiclass accuracy for the trained seen classifiers is $51.3\%$ and $15.4\%$, using the classeme features,  on Flower dataset and Birds dataset\footnote{Birds dataset is known to be a challenging dataset for fine-grained with engineered-features}, respectively. 
Table ~\ref{tbl:flowerbirdsmauauc} (middle part) shows the average $MAU$ metric over three seen/unseen splits for Flower dataset and one split on Birds dataset, respectively. 
Furthermore, the relative improvements of our  SVM-DT-kernel approach is reported against the baselines. On Flower dataset,  it is interesting to see that our approach achieved $9.1\%$ MAU, showing 182\% improvement over random guess, \ignore{, which is $17.7\%$ of the multi-class accuracy of the seen classes (i.e. $51.3\%$),} 
by predicting the unseen classifiers using just textual features as privileged information (i.e. $\mathcal{T}$ domain). It is important to mention that we achieved also $13.4\%$( $268\%$ improvement over random guess), in one of the splits (the 9.1\% is the average over 3 seen/unseen splits). Similarly on Birds dataset, we achieved $3.4\%$ MAU from text features, $132\%$  the random guess performance (further improved up to $224\%$ in next experiments). In addition to the unseen class performance, we report the performance on seen classes as an upper bound of zero-shot learning for both Flower (50.7\%) and Birds datasets (16\%). 
\ignore{, which is $22.7\%$ of the multi-class accuracy of the seen classes on the same dataset ($15.4\%$)}

\textit{AUC metric: }  Table~\ref{tbl:flowerbirdsmauauc} (bottom part)  shows the average AUC on the two datasets, compared to the baselines. More results and figures \ignore{Corresponding figures for Birds dataset } for our kernel approach are attached in the supplementary materials.

Looking at table~\ref{tbl:flowerbirdsmauauc},  notice that the proposed approach performs marginally similar to some baselines from AUC perspective. However, there is a clear improvement  in MAU  and Recall metrics. These results show the advantage of predicting classifiers in kernel space. Furthermore, the table shows that our SVM-DT-kernel approach outperforms our DT-kernel model. This indicates the advantage of the class separation, which is adjusted by the SVM-DT-kernel model. \ignore{In all these experiments, we used a setting of our SVM-DT-kernel model, where \small$C_{\beta}({ \textbf{T}} )\,$\normalsize is ignored (i.e. \small$\lambda_2 = 0$\normalsize)\ignore{; see Sec ~\ref{ss:tr}}). In order to study whether \small$C_{\beta}( \textbf{T} )\,$\normalsize is effective in unseen class prediction, we performed an extra experiment on Birds dataset, where \small$\lambda_2>0\,$\normalsize (e.g. \small$\lambda_2 =1$\normalsize). We found that MAU of our DT approach has slightly decreased (i.e from $2.95\&$  to $2.91\%$ ). Under the same setting, we also found that  \small$C_{\beta}( \textbf{T} )\,$\normalsize  slightly reduced the performance of SVM-DT from $9.1\%$ to $8.98\%$.MAU. This reflect our intuition argued in the approach section\ignore{Sec ~\ref{ss:tr}}. Hence, we suggest to assign \small$\lambda_2\,$\normalsize to $0$ for our purpose.} 

\subsubsection{Multiple Kernel Learning (MKL) Experiment}

This experiment shows the added value of  proposing a kernelized zero-shot learning approach. We conducted an experiment where the final kernel on the visual domain is produced by Multiple Kernel Learning \cite{MKKLAlgs11}. For the images, we extracted kernel descriptors for Birds dataset. Kernel descriptors provide a principled way to turn any pixel attribute to patch-level features, and are able to generate rich features from various recognition cues. We specifically used four types of kernels introduced by~\cite{bo_nips10} as follows: \textit{Gradient Match Kernels} that captures image variation based on predefined kernels on image gradients. \textit{Color Match Kernel} that describes patch appearance using two kernels on top of RGB and normalized RGB for regular images and intensity for grey images. These kernels capture image variation and visual apperances. For modeling the local shape, \textit{Local Binary Pattern} kernels have been applied. We computed these kernel descriptors on local image patches with fixed size 16 x 16 sampled densely over a grid with step size 8 in a spatial pyramid setting with four layers. The dense features are vectorized using codebooks of size 1000. This process ended up with a 120,000 dimensional feature for each image (30,000 for each type). Having extracted the four types of descriptors, we compute an rbf kernel matrix for each type separately. We learn the bandwidth parameters for each rbf kernel by cross validation on the seen classes. Then, we generate a new kernel \small$k_{mkl}(d, d') = \sum_{i=1}^4 w_i k_i(d, d')$\normalsize, such that $w_i$ is a weight assigned to each kernel. We learn these weights by applying Bucak's Multiple Kernel Learning algorithm \cite{nips10_Bucak}. Then, we applied our approach where the MKL-kernel is used in the visual domain and rbf kernel in the text domain similar to the previous experiments.

 To compare the kernel prediction approach to the linear prediction approach (formulation (E)) under this setting,  we concatenated  all kernel descriptors to form a 120,000 dimensional feature vector in the visual domain. As highlighted in  the kernel approach section, the linear prediction approach solves a quadratic program of \small$N+d_v+1\,$\normalsize variables for each unseen class.   Due to the large dimensionality of data  (\small$d_v = 120,000$\normalsize), this is not tractable. To make this setting applicable, we reduced the dimensionality of the feature vector into $4000$ using PCA\ignore{ to make it feasible to compute the performance}. This highlights the benefit of the kernelized approach since the quadratic program in Eq~\ref{eq:form_kernel} does not depend on the dimensionality of the feature space. Table~\ref{tbl:birdsmkl} shows MAU for the kernel prediction approaches under this setting against  linear prediction. The results show the benefit of zero-shot kernel prediction where an arbitrary kernel could be used to improve the performance.

\subsection{Multiple Representation Experiment and  Distributional Semantic(DS) Kernel}
\label{sec64}

This experiment elaborates the performance of kernel approach on different representations of text $\mathcal{T}$ and visual domains $\mathcal{V}$. 
In this experiment, we extracted Convolutional Neureal Network(CNN) image features for the Visual domain. We used caffe~\cite{jia2014caffe} implementation of~\cite{imagenetnips12}. Then, we extracted the sixth activation feature of the CNN (FC6) since we found it works the best on the standard classification setting. We found this consistent with the results of~\cite{donahue2014decaf} over different CNN layers. While using  TFIDF feature of text description and CNN features for images, we achieved 2.65\% for the linear version and 4.2\% for the rbf kernel on both text and images. We further improved the performance to 5.35\% by using our proposed Distributional Semantic (DS) kernel in the text domain and rbf kernel for images. In this DS experiment, we used the  distributional semantic model by~\cite{mikolov2013distributed} trained on  GoogleNews corpus (100 billion words)  resulting in a vocabulary of size 3 million words, and word vectors of $K=300$ dimensions. This experiment shows  the value of both the kernelized approach and the proposed kernel in Sec~\ref{dskernel}. We also applied the zero shot learning approach in~\cite{norouzi2014zero} which has the lowest performance in our settings; see Table~\ref{tbl:birdscnn}.


\textbf{Attributes Experiment: } Our  goal is not attribute prediction. However, it was interesting to see the behavior of our method where $\mathcal{T}$ space is defined from attributes instead of text. In contrast to attribute-based models, which fully utilize attribute information to build attribute classifiers, we do not learn attribute classifiers. In this experiment, our method  uses only the first moment of information of the attributes (i.e. the average attribute vector). We decided to compare to an attribute-based approach from this perspective. In particular, we applied the DAP attribute-based model~\cite{lampertPAMI13,Lampert09} to the Birds dataset, which is widely adopted in many applications (\textit{e.g.,} \cite{liu2013video,rohrbach11cvpr}). For visual domain, we used classeme features in this experiment (presented in  table~\ref{tbl:flowerbirdsmauauc} experiments.



An interesting result is that our approach achieved $5.6\%$  MAU ($224\%$ the random guess performance); see Table ~\ref{tbl:birds3}. In contrast, we get $4.8\%$ multiclass accuracy using  DAP approach~\cite{lampertPAMI13}. In this setting, we also measured the $N_{sc}$ to $ N_{sc}+1$ average recall. We found the recall measure is $76.7\%$ for our SVM-DT-kernel, while it is $68.1\%$ on  the DAP approach, which reflects better true positive rate (positive class is the unseen one). Most importantly, we achieved these results without learning any attribute classifiers, as in~\cite{lampertPAMI13}. When comparing the results  of our approach using attributes (Table~\ref{tbl:birds3}) vs. textual description (Table~\ref{tbl:flowerbirdsmauauc})\footnote{We are refering to the experiment that uses classeme as visual features to have a consistent comparison to here} as the $\mathcal{T}$ space used for prediction, it is clear that the attribute features give better predictions. This support our hypothesis that the more meaningful the \small$\mathcal{T}\,$\normalsize domain, the better the performance on \small$\mathcal{V}\,$\normalsize domain. This indicates that if a better textual representation is used, a better performance can be achieved. Attributes are good semantic representations of classes yet it is difficult to  define attributes for an arbitrary class and further measure the confidence of each one. In contrast, it is much easier to find an unstructured  text description for visual classes.

\vspace{-3mm}
\subsection{Experiments using deep image-sentence similarity and more recent Zero-shot learning methods}
\vspace{-1mm}
{We used a state of the art model~\cite{vendrov2016order} for image-sentence similarity by breaking down each text document into sentences and considering it as a positive sentence for all images in the corresponding class. Then we measure the similarities between an image to class by averaging its similarity to all sentences in that class. Images were encoded using VGGNet~\cite{simonyan2015very} and sentences were encoded by an RNN with GRU activations~\cite{cho2014learning}. The MAU on Birds dataset for this experiments resulted in 3.3\% MAU which is better that the Linear Classifier in Table~\ref{tbl:birdscnn}. However, our kernel method (Eq~\ref{eq:form_kernel}) over deep features is still performing 2.03\% better (i.e. 5.35\% MAU).

\subsection{SC-Seen Split on CUB 2011 ~\cite{akata2013label}}
\label{ss_soa}


We report the zero-shot performance on the SC-Seen (Super Category Seen) split, detailed in subsection A. We applied both our linear and kernel method and compare against  recently published results in our setting~\cite{qiao2016less,vendrov2016order,romera2015embarrassingly,akata2015evaluation}.  We performed all the experiments in the  previous sections (best zero-shot performance CUB dataset on SC-Unseen split  is 5.35\% on  SC-Unseen split designed in~\cite{Hoseini13}. It is not hard to see that the performance of our methods (both linear and kernel) on SC-Seen split is significantly better than SC-Unseen split designed in~\cite{Hoseini13}. Our kernel approach results on SC-Seen split is 33.5\% which is the best performing methods as shown in table~\ref{table_newsplit}.  When we used a binary version of Term Frequency (each word has 1 if exist, 0 otherwise), our performance is 26.5\%. This big performance gap  shows how challenging is SC-Unseen (Super Category Unseen) split compared to the SC-Seen split.  It is important to mention that the assumption of using existing images as negative examples is not valid on this split. Hence, we did not enforce this constraint on SC-Seen Split (constraints in Eq.~\ref{eq:form} for linear and in Eq.~\ref{eq:form_kernel} for the kernel version). Hence, the prediction is dominated by our Domain Transfer function. When we added these constraints our performance goes down from 33.5\% to 8\% which is expected due to the incorrectness of the assumption on this split. Based on this result, we encourage future researchers on this problem to report the performance on both SC-Unseen Split  and SC-Seen Split, where we showed that SC-Unseen Split is more challenging. 

\begin{table}
\caption{Zero-shot Learning Performance CUB Dataset (Super Category Seen (SC-Seen) Split)}
\label{table_newsplit}
\begin{tabular}{|c|c|}
\hline 
& MAU  \\ 
\hline 
\hline
{Our Kernel Classifier Prediction($\mathcal{V}$-rbf, $\mathcal{T}$-rbf on TFIDF)}& \textbf{33.5 \%} \\ 
{Our Kernel Classifier Prediction($\mathcal{V}$-rbf, $\mathcal{T}$-rbf on TFBin)}& \textbf{26.5 \%} \\ 
Our Linear Classifier Prediction using TFBin & \textbf{17.02 \% }  \\ 
\hline 
{Less is more ~\cite{qiao2016less}} CVPR 2016 & \textbf{29.0 \%} \\ 
Order Embedding~\cite{vendrov2016order} ICLR 2016 & 17.1 \%  \\ 
ESZSL~\cite{romera2015embarrassingly}  ICML 2016 & 23.8 \%  \\ 
Akata et al.~\cite{akata2015evaluation} CVPR 2015 with Word2vec & 23.8 \%  \\ 
Akata et al.~\cite{akata2015evaluation} CVPR 2015 with GloVE & 28.4 \%  \\ 
\hline 
\end{tabular}
\end{table}

}

\vspace{-2mm}
\section{Conclusion}
We explored the problem of predicting visual classifiers from textual description of classes with no training images.  We investigated and  experimented with different formulations for the problem within the fine-grained categorization context.  We first proposed  a novel formulation that captures information between the visual and textual domains by involving knowledge transfer from textual features to visual features, which indirectly leads to predicting a linear visual classifier described by the text. \ignore{In the future, we are planning to propose a kernel version to tackle the problem instead of using linear classifiers. Furthermore, } We also proposed a new zero-shot learning technique to predict kernel-classifiers of unseen categories using information from a privilege space. We formulated the problem as domain transfer function from text description  to the visual classification space, while supporting kernels in both domains. We proposed a one-class SVM adjustment to our domain transfer function in order to improve the prediction. We validated the performance of our model by several experiments. We illustrated the value of proposing a kernelized version by applying kernels generated by Multiple Kernel Learning (MKL) and achieved better results. 
 In the future, we aim to improve this model by learning the unseen classes jointly and on a larger scale. 

\ignore{
We are also looking forward to studying more features for the $\mathcal{X}$ and $\mathcal{E}$ domains in a  large scale setting (number of classes $>$ 1000). }

\noindent \textbf{Acknowledgment. } This research was funded by NSF award IIS-1409683 and  IIS-1218872. 


\ifCLASSOPTIONcaptionsoff
  \newpage
\fi

{
\bibliographystyle{IEEEtran}
\bibliography{egbib,write_a_classifier,elgammal,NLPVision,NLPVisionProposal,smara}
}
\vspace{-18mm}
\begin{IEEEbiography}[{\includegraphics[width=1.0in,height=1.0in,clip,keepaspectratio]{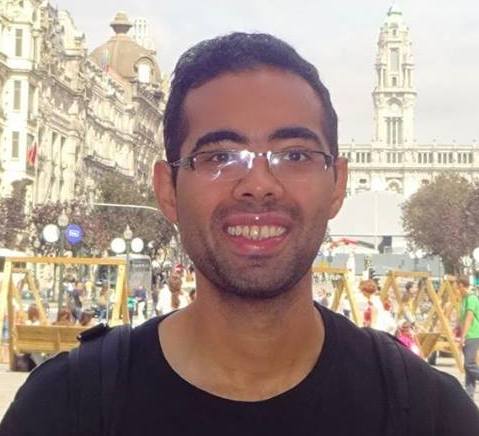}}]{Mohamed Elhoseiny}
 is a PostDoc Researcher at Facebook Research.
His primary research interest is in computer vision, machine learning, intersection between natural language and vision, language guided visual-perception,  and visual reasoning, art \& AI. He received his PhD degree from Rutgers University, New Brunswick, in 2016 under Prof. Ahmed Elgammal. Mohamed received an NSF Fellowship in 2014 for the Write-a-Classifier project (ICCV13), best intern award at SRI International 2014, and the Doctoral Consortium award at CVPR 2016.
\end{IEEEbiography}
\vspace{-17mm}
\begin{IEEEbiography}[\vspace{-4mm}{\includegraphics[width=1.0in,height=1.0in,clip,keepaspectratio]{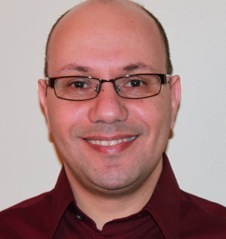}}]{Ahmed Elgammal} is a professor at the department
of computer science, Rutgers University.  His primary research interest is computer
vision and machine learning. His research focus
includes human motion analysis, tracking, human identification, and statistical
methods for computer vision. Dr. Elgammal
is a senior member of IEEE. Dr. Elgammal received his Ph.D. in
2002 from the University of Maryland, College Park.
\end{IEEEbiography}
\vspace{-20mm}
\begin{IEEEbiography}[\vspace{-5mm}{\includegraphics[width=0.9in,height=0.9in,clip,keepaspectratio]{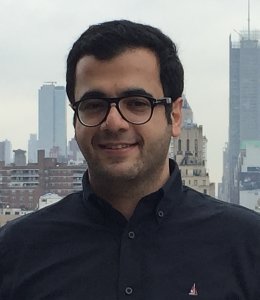}}]{Babak Saleh}
works on computer vision, machine learning, and human perception at Rutgers University. He is interested to build visual recognition systems with the human-level capability of image understanding. His line of research has been recognized by major media outlets and has received the outstanding student paper award from AAAI 2016. 
\end{IEEEbiography}
\vspace{-10mm}
\end{document}